\documentclass[10pt,journal,compsoc]{IEEEtran}

\ifCLASSOPTIONcompsoc
  \usepackage[nocompress]{cite}
\else
  \usepackage{cite}
\fi

\usepackage{ragged2e}
\usepackage{graphicx}
\usepackage{subcaption}
\usepackage{amsmath}
\usepackage{amssymb}
\usepackage{commath}
\usepackage{boldline}
\usepackage[symbol,perpage]{footmisc}
\usepackage{bm}

\newcommand{\nickname}{3D-RecGAN++}
\newcommand{\etal}{\textit{et al}.}
\newcommand{\ie}{\textit{i}.\textit{e}.,}
\newcommand{\eg}{\textit{e}.\textit{g}.,}
\newcommand{\etc}{\textit{etc}.}

\begin{document}
\title{Dense 3D Object Reconstruction \\ from a Single Depth View}

\author{Bo Yang,
        Stefano Rosa,
        Andrew Markham,
        Niki Trigoni,
        Hongkai Wen\textsuperscript{*}
\IEEEcompsocitemizethanks
{\IEEEcompsocthanksitem Bo Yang, Stefano Rosa, Andrew Markham and Niki Trigoni are with the Department of Computer Science, University of Oxford, UK.\protect\\
E-mail:\{bo.yang,stefano.rosa,andrew.markham,niki.trigoni\}@cs.ox.ac.uk
\IEEEcompsocthanksitem Corresponding to Hongkai Wen, who is with the Department of Computer Science, University of Warwick, UK.\protect\\
E-mail: hongkai.wen@dcs.warwick.ac.uk
}
}

\IEEEtitleabstractindextext{%
\begin{abstract}
\justifying
In this paper, we propose a novel approach, \textbf{\nickname{}}, which reconstructs the complete 3D structure of a given object from a single arbitrary depth view using generative adversarial networks. Unlike existing work which typically requires multiple views of the same object or class labels to recover the full 3D geometry, the proposed \nickname{} only takes the voxel grid representation of a depth view of the object as input, and is able to generate the complete 3D occupancy grid with \textbf{a high resolution of} $\boldsymbol{256^3}$ by recovering the occluded/missing regions. The key idea is to combine the generative capabilities of 3D encoder-decoder and the conditional adversarial networks framework, to infer accurate and fine-grained 3D structures of objects in high-dimensional voxel space. Extensive experiments on large synthetic datasets and real-world Kinect datasets show that the proposed \nickname{} significantly outperforms the state of the art in single view 3D object reconstruction, and is able to reconstruct unseen types of objects.
\end{abstract}

\begin{IEEEkeywords}
3D Reconstruction, Shape Completion, Shape inpainting, Single Depth View, Adversarial Learning, Conditional GAN.
\end{IEEEkeywords}}

\maketitle

\IEEEdisplaynontitleabstractindextext
\IEEEpeerreviewmaketitle
\IEEEraisesectionheading{\section{Introduction}\label{sec:intro}}
\IEEEPARstart{T}{o} reconstruct the complete and precise 3D geometry of an object is essential for many graphics and robotics applications, from AR/VR \cite{B2016e} and semantic understanding, to object deformation \cite{Wang2018a}, robot grasping \cite{Varley2017} and obstacle avoidance. Classic approaches use the off-the-shelf low-cost depth sensing devices such as Kinect and RealSense cameras to recover the 3D shape of an object from captured depth images. Those approaches typically require multiple depth images from different viewing angles of an object to estimate the complete 3D structure \cite{Newcombe2011a}\cite{Niener2013}\cite{Steinbr2013}. However, in practice it is not always feasible to scan all surfaces of an object before reconstruction, which leads to incomplete 3D shapes with occluded regions and large holes. In addition, acquiring and processing multiple depth views require more computing power, which is not ideal in many applications that require real-time performance.

We aim to tackle the problem of estimating the complete 3D structure of an object using a single depth view. This is a very challenging task, since the partial observation of the object (\ie{} a depth image from one viewing angle) can be theoretically associated with an infinite number of possible 3D models. Traditional reconstruction approaches typically use interpolation techniques such as plane fitting, Laplacian hole filling \cite{Nealen2006}\cite{Zhao2007}, or Poisson surface estimation \cite{Kazhdan2006}\cite{Kazhdan2013} to infer the underlying 3D structure. However, they can only recover very limited occluded or missing regions, \eg{} small holes or gaps due to quantization artifacts, sensor noise and insufficient geometry information.

Interestingly, humans are surprisingly good at solving such ambiguity by implicitly leveraging prior knowledge. For example, given a view of a chair with two rear legs occluded by front legs, humans are easily able to guess the most likely shape behind the visible parts. Recent advances in deep neural networks and data driven approaches show promising results in dealing with such a task.

In this paper, we aim to acquire the complete and high-resolution 3D shape of an object given a single depth view. By leveraging the high performance of 3D convolutional neural nets and large open datasets of 3D models, our approach learns a smooth function that maps a 2.5D view to a complete and dense 3D shape. In particular, we train an end-to-end model which estimates full volumetric occupancy from a single 2.5D depth view of an object.

While state-of-the-art deep learning approaches \cite{Wu2015}\cite{Dai2017b}\cite{Varley2017} for 3D shape reconstruction from a single depth view achieve encouraging results, they are limited to very small resolutions, typically at the scale of $32^3$ voxel grids. As a result, the learnt 3D structure tends to be coarse and inaccurate. In order to generate higher resolution 3D objects with efficient computation, Octree representation has been recently introduced in \cite{Tatarchenko2017}\cite{Riegler2017}\cite{Christian2017}. However, increasing the density of output 3D shapes would also inevitably pose a great challenge to learn the geometric details for high resolution 3D structures, which has yet to be explored.

Recently, deep generative models achieve impressive success in modeling complex high-dimensional data distributions, among which Generative Adversarial Networks (GANs) \cite{Goodfellow2014} and Variational Autoencoders (VAEs) \cite{Kingma2014} emerge as two powerful frameworks for generative learning, including image and text generation \cite{Hu2017a}\cite{Karras2017}, and latent space learning \cite{Chen2016a}\cite{Kulkarni2015}. In the past few years, a number of works \cite{B2016f}\cite{Girdhar}\cite{Huang2015a}\cite{Wu2016a} applied such generative models to learn latent space to represent 3D object shapes, in order to solve tasks such as new image generation, object classification, recognition and shape retrieval.

In this paper, we propose \nickname{}, a simple yet effective model that combines a skip-connected 3D encoder-decoder with adversarial learning to generate a complete and fine-grained 3D structure conditioned on a single 2.5D view. Particularly, our model firstly encodes the 2.5D view to a compressed latent representation which implicitly represents general 3D geometric structures, then decodes it back to the most likely full 3D shape. Skip-connections are applied between the encoder and decoder to preserve high frequency information. The rough 3D shape is then fed into a conditional discriminator which is adversarially trained to distinguish whether the coarse 3D structure is plausible or not. The encoder-decoder is able to approximate the corresponding shape, while the adversarial training tends to add fine details to the estimated shape. To ensure the final generated 3D shape corresponds to the input single partial 2.5D view, adversarial training of our model is based on a conditional GAN \cite{Mirza2014} instead of random guessing. The above network excels the competing approaches \cite{Varley2017}\cite{Dai2017b}\cite{Han2017}, which either use a single fully connected layer \cite{Varley2017}, a low capacity decoder without adversarial learning \cite{Dai2017b}, or the multi-stage and ineffective LSTMs \cite{Han2017} to estimate the full 3D shapes.

Our contributions are as follows:

(1) We propose a simple yet effective model to reconstruct the complete and accurate 3D structure using a single arbitrary depth view. Particularly, our model takes a simple occupancy grid map as input without requiring object class labels or any annotations, while predicting a compelling shape within a high resolution of \textbf{ $256^3$ } voxel grid. By drawing on both 3D encoder-decoder and adversarial learning, our approach is end-to-end trainable with high level of generality. 

(2) We exploit conditional adversarial training to refine the 3D shape estimated by the encoder-decoder. Our contribution here is that we use the mean value of a latent vector feature, instead of a single scalar, as the output of the discriminator to stabilize GAN training. 

(3) We conduct extensive experiments for single category and multi-category object reconstruction, outperforming the state of the art. Importantly, our approach is also able to generalize to previously unseen object categories. At last, our model also performances robustly on real-world data, after being trained purely on synthetic datasets.

(4) To the best of our knowledge, there are no good open datasets which have the ground truth for occluded/missing parts and holes for each 2.5D view in real-world scenarios. We therefore collect and release our real-world testing dataset to the community.

A preliminary version of this work has been published in ICCV 2017  workshops \cite{Yang2017b}. Our code and data are available at:\textit{ https://github.com/Yang7879/3D-RecGAN-extended}

\section{Related Work}\label{sec:liter}
We review different pipelines for 3D reconstruction or shape completion. Both conventional geometry based techniques and the state of the art deep learning approaches are covered.

(1) \textbf{3D Model/Shape Completion.} Monszpart \etal{} use plane fitting to complete small missing regions in \cite{Monszpart2015a}, while shape symmetry is applied in \cite{Mitra2006}\cite{Pauly2008}\cite{Sipiran2014}\cite{Speciale2016}\cite{Thrun2005} to fill in holes. Although these methods show good results, relying on predefined geometric regularities fundamentally limits the structure space to hand-crafted shapes. Besides, these approaches are likely to fail when missing or occluded regions are relatively big. Another similar fitting pipeline is to leverage database priors. Given a partial shape input, an identical or most likely 3D model is retrieved and aligned with the partial scan in \cite{Kim2012}\cite{Li2015a}\cite{Nan}\cite{Shao2012}\cite{Shi2016}\cite{Rock2015}. However, these approaches explicitly assume the database contains identical or very similar shapes, thus being unable to generalize to novel objects or categories.

(2) \textbf{Multiple RGB/Depth Images Reconstruction.} Traditionally, 3D dense reconstruction in SfM and visual SLAM requires a collection of RGB images \cite{Hartley2004}. Geometric shape is recovered by dense feature extraction and matching \cite{Newcombe2011b}, or by directly minimizing reprojection errors \cite{Baker2004} from color images. Shape priors are also concurrently leveraged with the traditional multi-view reconstruction for dense object shape estimation in \cite{Bao2013}\cite{Dame2013}\cite{Engelmann2016}. Recently, deep neural nets are designed to learn the 3D shape from multiple RGB images in \cite{Chan2016}\cite{B2016h}\cite{Lun2017}\cite{Rezende2016}\cite{Kar2017}\cite{Ji2017b}. However, resolution of the recovered occupancy shape is usually up to a small scale of $32^3$. With the advancement of depth sensors, depth images are also used to recover the object shape. Classic approaches usually fuse multiple depth images through iterative closest point (ICP) algorithms \cite{Newcombe2011a}\cite{Whelan2012}\cite{Whelan2015}, while recent work \cite{Riegler2017} learns the 3D shape using deep neural nets from multiple depth views.

(3) \textbf{Single RGB Image Reconstruction.} Predicting a complete 3D object model from a single view is a long-standing and extremely challenging task. When reconstructing a specific object category, model templates can be used. For example, morphable 3D models are exploited for face recovery \cite{Blanz2003}\cite{Dou2017}. This concept was extended to reconstruct simple objects in \cite{Kar2015a}. For general and complex object reconstruction from a single RGB image, recent works \cite{Gwak2017}\cite{Tulsiani2017}\cite{Yan2016} aim to infer 3D shapes using multiple RGB images for weak supervision. Shape prior knowledge is utilized in \cite{Kong2017}\cite{Kurenkov2017}\cite{Murthy2016} for shape estimation. To recover high resolution 3D shapes, Octree representation is introduced in \cite{Tatarchenko2017}\cite{Riegler2017}\cite{Christian2017} to save computation, while an inverse discrete cosine transform (IDCT) technique is proposed in \cite{Johnston2017}. Lin \etal{}\cite{Lin2017a} designed a pseudo-renderer to predict dense 3D shapes, while 2.5D sketches and dense 3D shapes are sequentially estimated from a single RGB image in \cite{Wu2017d}.

(4) \textbf{Single Depth View Reconstruction.} The task of reconstruction from a single depth view is to complete the occluded 3D structures behind the visible parts. 3D ShapeNets \cite{Wu2015} is among the early work using deep neural nets to estimate 3D shapes from a single depth view. Firman \etal{} \cite{Firman2016} trained a random decision forest to infer unknown voxels. Originally designed for shape denoising, VConv-DAE \cite{B2016e} can also be used for shape completion. To facilitate robotic grasping, Varley \etal{} proposed a neural network to infer the full 3D shape from a single depth view in \cite{Varley2017}. However, all these approaches are only able to generate low resolution voxel grids which are less than $40^3$ and unlikely to capture fine geometric details. Recent works \cite{Dai2017b} \cite{Song2017}\cite{Han2017}\cite{Wang2017b} can infer higher resolution 3D shapes. However, the pipeline in \cite{Dai2017b} relies on a shape database to synthesize a higher resolution shape after learning a small $32^3$ voxel grid from a depth view, while SSCNet \cite{Song2017} requires voxel-level annotations for supervised scene completion and semantic label prediction.
Both \cite{Han2017} and \cite{Wang2017b} were originally designed for shape inpainting instead of directly reconstructing the complete 3D structure from a partial depth view. The recent 3D-PRNN \cite{Zou2017} predicts simple shape primitives using RNNs, but the estimated shapes do not have finer geometric details.

(5) \textbf{Deep Generative Frameworks.} Deep generative frameworks, such as VAEs \cite{Kingma2014} and GANs \cite{Goodfellow2014}, have achieved impressive success in image super-resolution \cite{Ledig2016}, image generation \cite{Karras2017}, text to image synthesis \cite{Reed2016}, \etc{} VAE and GAN are further combined in \cite{Larsen2016} and achieve compelling results in learning visual features. Recently, generative networks are applied in \cite{Gadelha2016}\cite{Smith2017}\cite{Soltani2017}\cite{Wu2016a} to generate low resolution 3D structures. However, incorporating generative adversarial learning to estimate high resolution 3D shapes is not straightforward, as it is difficult to generate samples for high dimensional and complex data distributions \cite{Arjovsky2017} and this may lead to the instability of adversarial generation.

\section{\nickname{}}
\subsection{Overview}\label{sec:overview}
\begin{figure}[t]
	\setlength{\abovecaptionskip}{ 0 cm}
    \setlength{\belowcaptionskip}{ -8pt}
    \centering
    \includegraphics[width=0.45\textwidth]{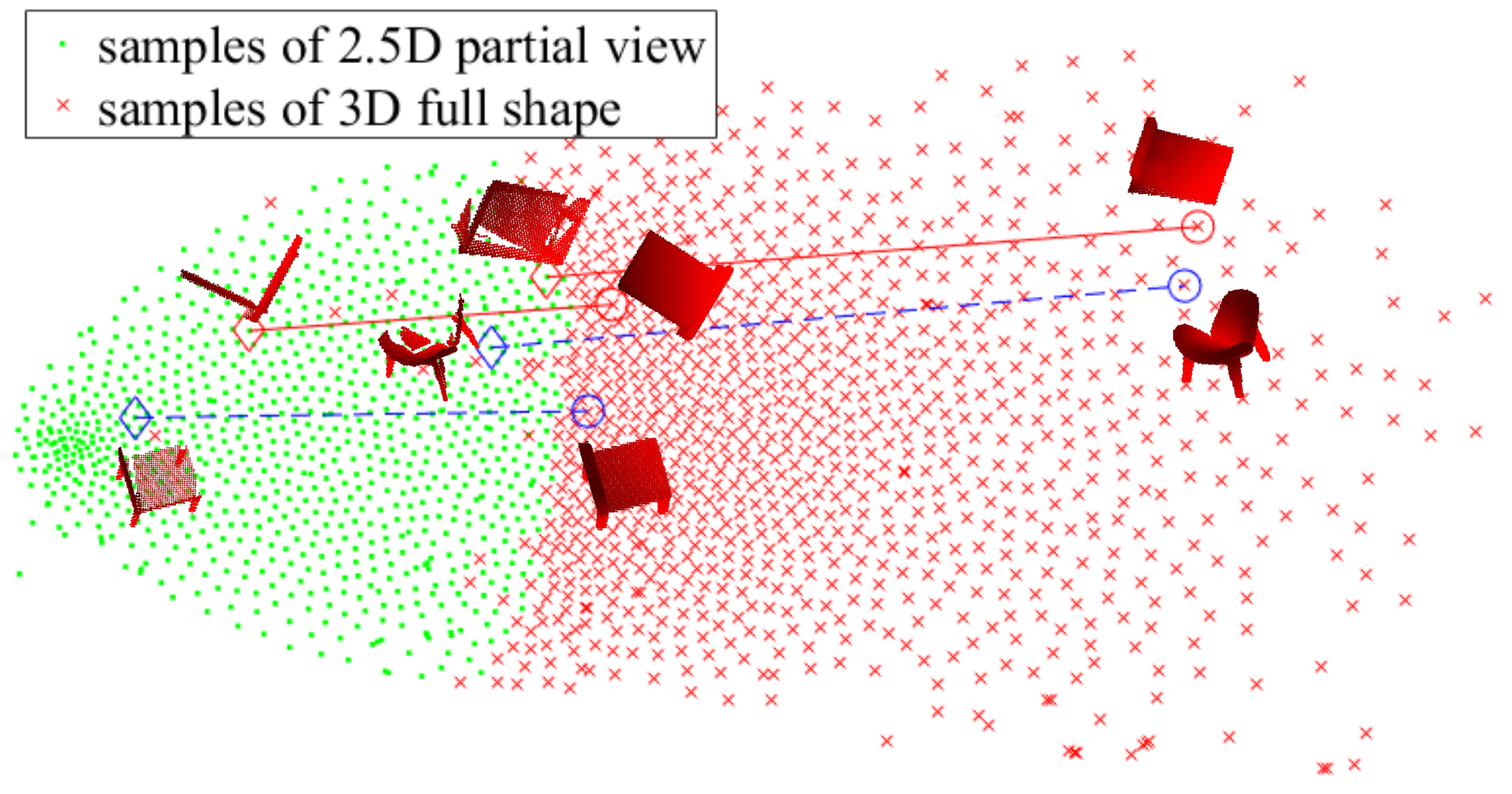}
    \caption{t-SNE embeddings of 2.5D partial views and 3D complete shapes of multiple object categories.}
    \label{fig:25d_3d}
    \vspace{-0.3 cm}
\end{figure}

Our method aims to estimate a complete and dense 3D structure of an object, which only takes an arbitrary single 2.5D depth view as input. The output 3D shape is automatically aligned with the corresponding 2.5D partial view. To achieve this task, each object model is represented by a high resolution 3D voxel grid. We use the simple occupancy grid for shape encoding, where $1$ represents an occupied cell and $0$ an empty cell. Specifically, the input 2.5D partial view, denoted as $\boldsymbol{x}$, is a $64^3$ occupancy grid, while the output 3D shape, denoted as $\boldsymbol{y}$, is a high resolution $256^3$ probabilistic voxel grid. The input partial shape is directly calculated from a single depth image given camera parameters. We use the ground truth dense 3D shape with aligned orientation as same as the input partial 2.5D depth view to supervise our network.

\begin{figure}[t]
	\setlength{\abovecaptionskip}{ 0 cm}
    \setlength{\belowcaptionskip}{ -8pt}
    \centering
    \includegraphics[width=0.48\textwidth]{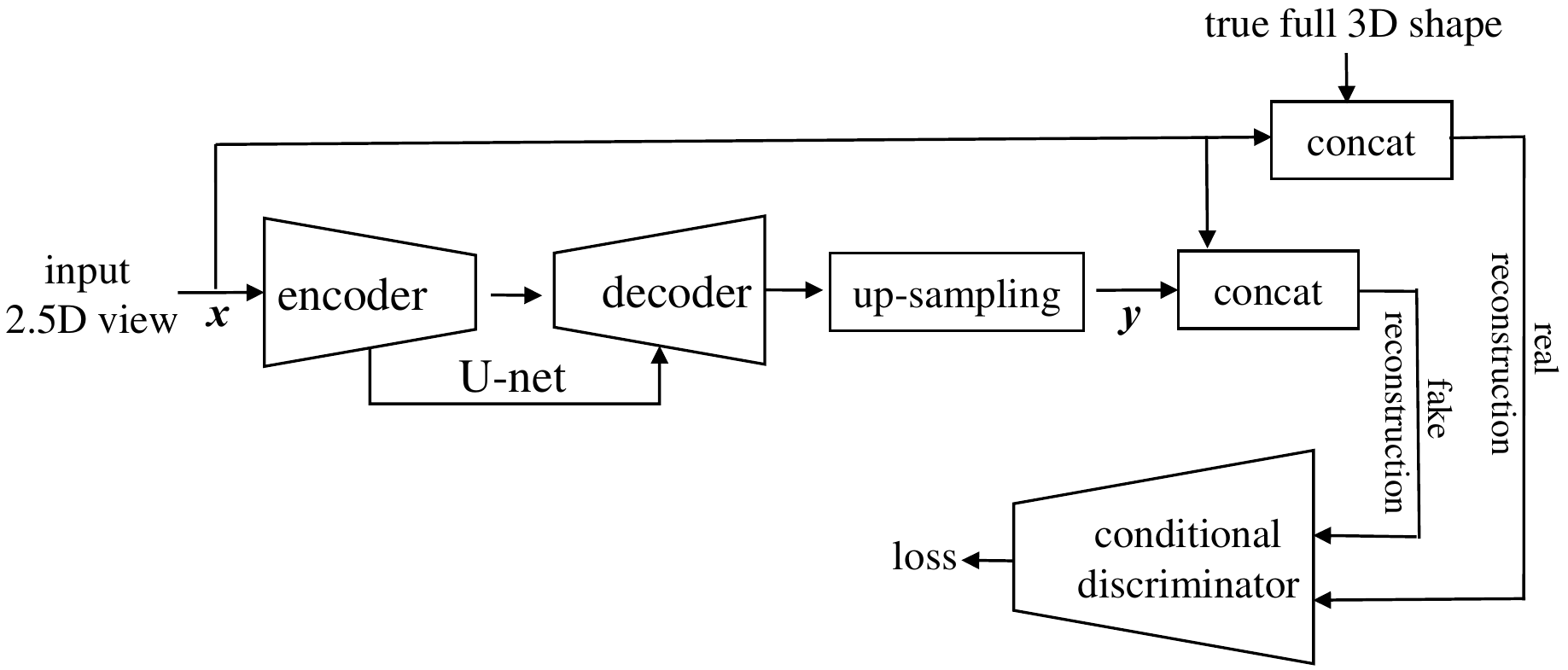}
    \caption{Overview of the network architecture for training.}
    \label{fig:arch_train}
\end{figure}
\begin{figure}[t]
	\vspace{-0.1 cm}
	\setlength{\abovecaptionskip}{ 0 cm}
    \setlength{\belowcaptionskip}{ -8pt}
    \centering
    \includegraphics[width=0.48\textwidth]{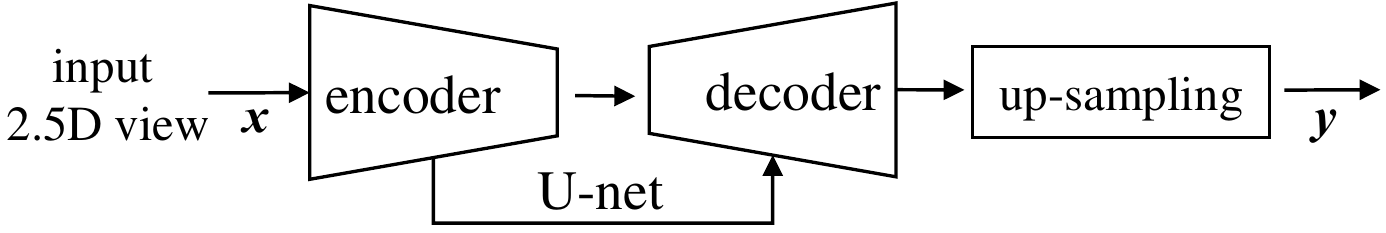}
    \caption{Overview of the network architecture for testing.}
    \label{fig:arch_test}
    \vspace{-0.3 cm}
\end{figure}

To generate ground truth training and evaluation pairs, we virtually scan 3D objects from ShapeNet \cite{Chang2015}. Figure \ref{fig:25d_3d} is the t-SNE visualization \cite{Maaten2008} of partial 2.5D views and the corresponding full 3D shapes for multiple general chair and bed models. Each green dot represents the t-SNE embedding of a 2.5D view, whilst a red dot is the embedding of the corresponding 3D shape. It can be seen that multiple categories inherently have similar 2.5D to 3D mapping relationships. Essentially, our neural network is to learn a smooth function, denoted as $f$, which maps green dots to red dots as close as possible in high dimensional space as shown in Equation \ref{eq:f25d3d}. The function $f$ is parametrized by neural layers in general.
\vspace{-0.1 cm}
\begin{equation}
\label{eq:f25d3d}
\boldsymbol{y} = f(\boldsymbol{x}) \quad \left( \boldsymbol{x} \in Z^{64^3}, where {\ } Z=\{0,1\} \right)
\end{equation}

After generating training pairs, we feed them into our network. The first part of our network loosely follows the idea of a 3D encoder-decoder with the U-net connections \cite{Ronneberger2015}. The skip-connected encoder-decoder serves as an initial coarse generator which is followed by an up-sampling module to further generate a higher resolution 3D shape within a $256^3$ voxel grid. This whole generator learns a correlation between partial and complete 3D structures. With the supervision of complete 3D labels, the generator is able to learn a function $f$ and infer a reasonable 3D shape given a brand new partial 2.5D view. During testing, however, the results tend to be grainy and without fine details.

To address this issue, in the training phase, the reconstructed 3D shape from the generator is further fed into a conditional discriminator to verify its plausibility. In particular, a partial 2.5D input view is paired with its corresponding complete 3D shape, which is called the `real reconstruction', while the partial 2.5D view is paired with its corresponding output 3D shape from generator, which is called the `fake reconstruction'. The discriminator aims to discriminate all `fake reconstruction' from `real reconstruction'. In the original GAN framework \cite{Goodfellow2014}, the task of the discriminator is to simply classify real and fake inputs, but its Jensen-Shannon divergence-based loss function is difficult to converge. The recent WGAN \cite{Arjovsky2017} leverages Wasserstein distance with weight clipping as a loss function to stabilize the training procedure, whilst the extended work WGAN-GP \cite{Gulrajani2017} further improves the training process using a gradient penalty with respect to its input. In our \nickname{}, we apply WGAN-GP as the loss function on top of the mean feature of our conditional discriminator, which guarantees fast and stable convergence. The overall network architecture for training is shown in Figure \ref{fig:arch_train}, while the testing phase only needs the well trained generator as shown in Figure \ref{fig:arch_test}. 

Overall, the main challenge of 3D reconstruction from an arbitrary single view is to generate new information including filling the missing and occluded regions from unseen views, while keeping the estimated 3D shape corresponding to the specific input 2.5D view. In the training phase, our \nickname{} firstly leverages a skip-connected encoder-decoder together with an up-sampling module to generate a reasonable `fake reconstruction' within a high resolution occupancy grid, then applies adversarial learning to refine the `fake reconstruction' to make it as similar to `real reconstruction' by jointly updating parameters of the generator. In the testing phase, given a novel 2.5D view as input, the jointly trained generator is able to recover a full 3D shape with satisfactory accuracy, while the discriminator is no longer used. 
\begin{figure*}[h]
\setlength{\abovecaptionskip}{ 0 cm}
\setlength{\belowcaptionskip}{ 4 pt}
\centering
\begin{subfigure}[t]{0.98\textwidth}
   \includegraphics[width=1\linewidth]{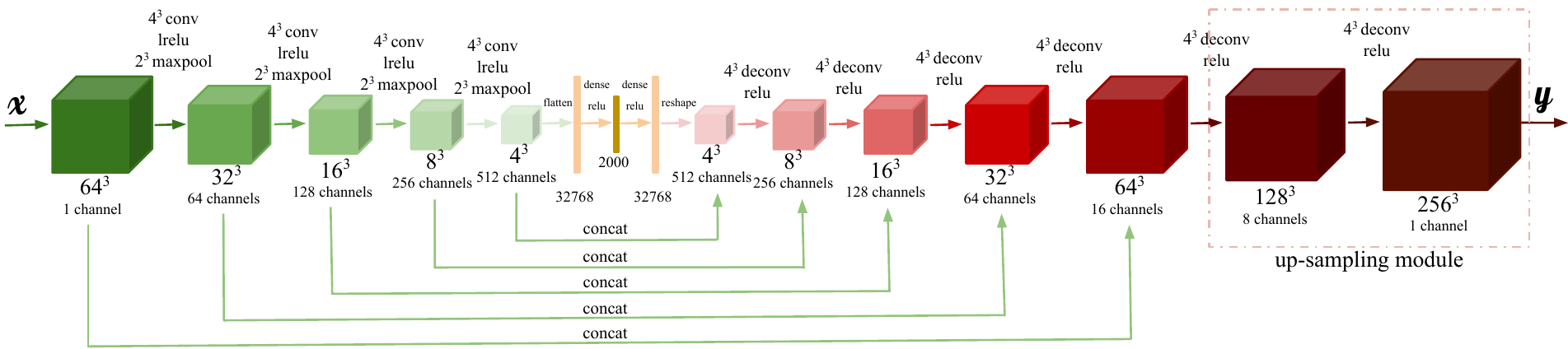}
   \caption{Generator for 3D shape estimation from a single depth view.}
   \label{fig:gen} 
\end{subfigure}
\begin{subfigure}[t]{0.98\textwidth}
   \includegraphics[width=1\linewidth]{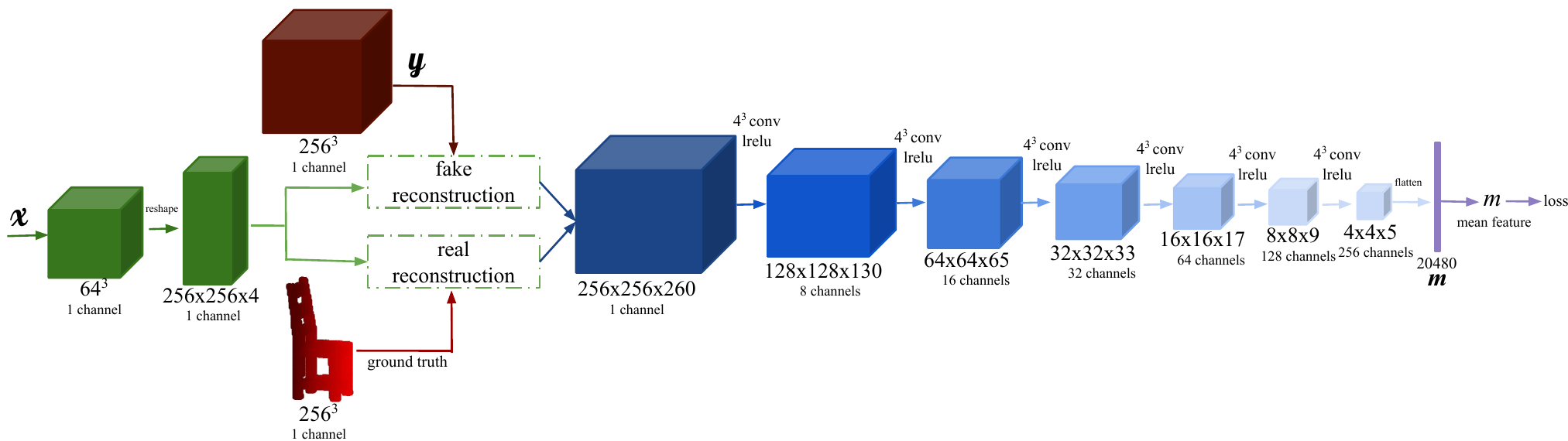}
   \caption{Discriminator for 3D shape refinement.}
   \label{fig:dis}
\end{subfigure}
\caption{Detailed architecture of \nickname{}, showing the two main building blocks. Note that, although these are shown as two separate modules, they are trained end-to-end.}
\label{fig:net_detail}
\vspace{-0.6 cm}
\end{figure*}

\subsection{Architecture}
Figure \ref{fig:net_detail} shows the detailed architecture of our proposed \nickname{}. It consists of two main networks: the generator as in Figure \ref{fig:gen} and the discriminator as in Figure \ref{fig:dis}.

\textbf{The generator} consists of a skip-connected encoder-decoder and an up-sampling module. Unlike the vanilla GAN generator which generates data from arbitrary latent distributions, our \nickname{} generator synthesizes data from 2.5D views. Particularly, the encoder has five 3D convolutional layers, each of which has a bank of $4\times4\times4$ filters with strides of $1\times1\times1$, followed by a leaky ReLU activation function and a max pooling layer with $2\times2\times2$ filters and strides of $2\times2\times2$. The number of output channels of max pooling layer starts with 64, doubling at each subsequent layer and ends up with 512. The encoder is lastly followed by two fully-connected layers to embed semantic information into a latent space. The decoder is composed of five symmetric up-convolutional layers which are followed by ReLU activations. Skip-connections between encoder and decoder guarantee propagation of local structures of the input 2.5D view. The skip-connected encoder-decoder is followed by the up-sampling module which simply consists of two layers of up-convolutional layers as detailed in Figure \ref{fig:gen}. This simple up-sampling module directly upgrades the output 3D shape to a higher resolution of $256^3$ without requiring complex network design and operations. It should be noted that without the two fully connected layers and skip-connections, the vanilla encoder-decoder would be unable to learn reasonable complete 3D structures as the latent space is limited and the local structure is not preserved. 
The loss function and optimization methods are described in Section \ref{sec:loss}.

\textbf{The discriminator} aims to distinguish whether the estimated 3D shapes are plausible or not. Based on the conditional GAN, the discriminator takes both real reconstruction pairs and fake reconstruction pairs as input. In particular, it consists of six 3D convolutional layers, the first of which concatenates the generated 3D shape (\ie{} a $256^3$ voxel grid) and the input 2.5D partial view (\ie{} a $64^3$ voxel grid), reshaped as a $256\times256\times4$ tensor. The reshaping process is done straightforwardly using Tensorflow `tf.reshape()'. Basically, this is to inject the condition information with a matched tensor dimension, and then leave the network itself to learn useful features from this condition input. Each convolutional layer has a bank of $4\times4\times4$ filters with strides of $2\times2\times2$, followed by a ReLU activation function except for the last layer which is followed by a sigmoid activation function. The number of output channels of the convolutional layers starts with 8, doubling at each subsequent layer and ends up with 256. The output of the last neural layer is reshaped as a latent vector which is the latent feature of discriminator, denoted as $\bm{m}$.

\subsection{Mean Feature for Discriminator}
At the early training stage of GAN, as the high dimensional real and fake distributions may not overlap, the discriminator can separate them perfectly using a single scalar output, which is analyzed in \cite{Arjovsky2017c}. In our experiments, due to the extremely high dimensionality (\ie{} $256^3+64^3$ dimensions) of the input data pair, the WGAN-GP always crashes in the early 3 epochs if we use a standard fully-connected layer followed by a single scalar as the final output for the discriminator. 

To stabilize training, we propose to use the mean feature $m$ (\ie{} mean of a vector feature $\bm{m}$) for discrimination. As the mean vector feature tends to capture more information from the input overall, it is more difficult for the discriminator to easily distinguish between fake or real inputs. This enables useful information to back-propagate to the generator. The final output of the discriminator  $D()$ is defined as:
\vspace{-0.1 cm}
\begin{equation}
\label{eq:mean_feature}
	m = \mathbf{E}(\bm{m})
\end{equation}

Mean feature matching is also studied and applied in \cite{Bao2017b}\cite{Mroueh2017b} to stabilize GAN. However, Bao \etal{} \cite{Bao2017b} minimize the $L_2$ loss of the mean feature, as well as the original Jensen-Shannon divergence-based loss \cite{Goodfellow2014}, requiring hyper-parameter tuning to balance the two losses. By comparison, in our \nickname{} setting, the mean feature of discriminator is directly followed by the existing WGAN-GP loss, which is simple yet effective to stabilize the adversarial training.

Overall, our discriminator learns to distinguish the distributions of mean feature of fake and real reconstructions, while the generator is trained to make the two mean feature distributions as similar as possible.

\subsection{Objectives}\label{sec:loss}
The objective function of \nickname{} includes two main parts: an object reconstruction loss $\ell_{en}$ for the generator; the objective function $\ell_{gan}$ for the conditional GAN.

(1) \textbf{$\ell_{en}$} For the generator, inspired by \cite{Brock2016}, we use modified binary cross-entropy loss function instead of the standard version. The standard binary cross-entropy weights both false positive and false negative results equally. However, most of the voxel grid tends to be empty, so the network easily gets a false positive estimation. In this regard, we impose a higher penalty on false positive results than on false negatives. Particularly, a weight hyper-parameter $\alpha$ is assigned to false positives, with (1-$\alpha$) for false negative results, as shown in  Equation \ref{eq:lae}.
\begin{equation}
\label{eq:lae}
\ell_{en} = \frac{1}{N} \sum_{i=1}^{N} \Big[-\alpha  \bar{y_i} \log(y_i) - (1-{\alpha} ) (1-\bar{y_i}) \log(1-y_i) \Big]
\end{equation}
$where$ $\bar{y_i}$ is the target value \{0,1\} of a specific $i^{th}$ voxel in the ground truth voxel grid $\boldsymbol{\bar{y}}$, and $y_i$ is the corresponding estimated value (0,1) in the same voxel from the generator output $\boldsymbol{y}$. We calculate the mean loss over the total $N$ voxels in the whole voxel grid.

(2) \textbf{$\ell_{gan}$} For the discriminator, we leverage the state of the art WGAN-GP loss functions. Unlike the original GAN loss function which presents an overall loss for both real and fake inputs, we separately represent the loss function $\ell_{gan}^{g}$ in Equation \ref{eq:lgang} for generating fake reconstruction pairs and $\ell_{gan}^{d}$ in Equation \ref{eq:lgand} for discriminating fake and real reconstruction pairs. Detailed definitions and derivation of the loss functions can be found in \cite{Arjovsky2017}\cite{Gulrajani2017}, but we modify them for our conditional GAN settings.
\begin{equation}
\label{eq:lgang}
\ell_{gan}^{g} = -\mathbf{E}\left[ D(\boldsymbol{y}|\boldsymbol{x}) \right]
\end{equation}
\vspace{-.5 cm}
\begin{align}
\label{eq:lgand}
\ell_{gan}^{d} = \mathbf{E}\left[ D(\boldsymbol{y}|\boldsymbol{x})\right] - \mathbf{E} \left[ D(\boldsymbol{\bar{y}}|\boldsymbol{x}) \right] \nonumber \\
+ \lambda \mathbf{E} \left[ \left(\norm{\nabla_{\boldsymbol{\hat{y}} }D(\boldsymbol{\hat{y}}|\boldsymbol{x})}_2 - 1 \right)^2  \right]
\end{align}
$where$ $\boldsymbol{\hat{y}} = \epsilon \boldsymbol{\bar{y}} +(1-\epsilon)\boldsymbol{y}, \epsilon \sim U[0,1]$, $\boldsymbol{x}$ is the input partial depth view, $\boldsymbol{y}$ is the corresponding output of the generator, $\boldsymbol{\bar{y}}$ is the corresponding ground truth. $\lambda$ controls the trade-off between optimizing the gradient penalty and the original objective in WGAN.

For the generator in our \nickname{} network, there are two loss functions, $\ell_{en}$ and $\ell_{gan}^{g}$, to optimize. As we discussed in Section \ref{sec:overview}, minimizing $\ell_{en}$ tends to learn the overall 3D shapes, whilst minimizing $\ell_{gan}^{g}$ estimates more plausible 3D structures conditioned on input 2.5D views. To minimize $\ell_{gan}^{d}$ is to improve the performance of discriminator to distinguish fake and real reconstruction pairs. To jointly optimize the generator, we assign weights $\beta$ to $\ell_{en}$ and $(1-\beta)$ to $\ell_{gan}^{g}$. Overall, the loss functions for generator and discriminator are as follows:
\begin{equation}
\ell_{g} = \beta \ell_{en} + (1-\beta) \ell_{gan}^{g}
\end{equation}
\begin{equation}
\ell_d = \ell_{gan}^{d}
\end{equation}

\subsection{Training}
We adopt an end-to-end training procedure for the whole network. To simultaneously optimize both generator and discriminator, we alternate between one gradient decent step on the discriminator and then one step on the generator. For the WGAN-GP, $\lambda$ is set as 10 for gradient penalty as in \cite{Gulrajani2017}. $\alpha$ ends up as 0.85 for our modified cross entropy loss function, while $\beta$ is 0.2 for the joint loss function $\ell_{g}$.

The Adam solver\cite{Kingma2015a} is used for both discriminator and generator with a batch size of 4. The other three Adam parameters are set to default values. Learning rate is set to $5e^{-5}$ for the discriminator and $1e^{-4}$ for the generator in all epochs. As we do not use dropout or batch normalization, the testing phase is exactly the same as the training stage. The whole network is trained on a single Titan X GPU from scratch.

\subsection{Data Synthesis}
For the task of 3D dense reconstruction from a single depth view, obtaining a large amount of training data is an obstacle. Existing real RGB-D datasets for surface reconstruction suffer from occlusions and missing data and there is no ground truth of complete and high resolution $256^3$ 3D shapes for each view. The recent work \cite{Dai2017b} synthesizes data for 3D object completion, but the object resolution is only up to $128^3$. 

To tackle this issue, we use the ShapeNet \cite{Chang2015} database to generate a large amount of training and testing data with synthetically rendered depth images and the corresponding complete 3D shape ground truth. Interior parts of individual objects are set to be filled, \ie{} `1', while the exterior to be empty, \ie{} `0'. A subset of object categories and CAD models are selected for our experiments. As some CAD models in ShapeNet may not be watertight, in our ray tracing based voxelization algorithm, if a specific point is inside of more than 5 faces along X, Y and Z axes, that point is deemed to be interior of the object and set as `1', otherwise `0'.

For each category, to generate \textbf{training data}, around 220 CAD models are randomly selected. For each CAD model, we create a virtual depth camera to scan it from 125 different viewing angles, 5 uniformly sampled views for each of roll, pitch and yaw space ranging from $0\sim2\pi$ individually. Note that, the viewing angles for all 3D models are the same for simplicity. For each virtual scan, both a depth image and the corresponding complete 3D voxelized structure are generated with regard to the same camera angle. That depth image is simultaneously transformed to a point cloud using virtual camera parameters \cite{Khoshelham2012} followed by voxelization which generates a partial 2.5D voxel grid. Then a pair of partial 2.5D view and the complete 3D shape is synthesized. Overall, around 26K training pairs are generated for each 3D object category.

For each category, to synthesize \textbf{testing data}, around 40 CAD models are randomly selected. For each CAD model, two groups of testing data are generated. \textbf{Group 1}, each model is virtually scanned from 125 viewing angles which are the same as used in training dataset. Around 4.5k testing pairs are generated in total. This group of testing dataset is denoted as \textbf{same viewing} (SV) angles testing dataset. \textbf{Group 2}, each model is virtually scanned from 216 different viewing angles, 6 uniformly sampled views from each of roll, pitch and yaw space ranging from $0\sim2\pi$ individually. Note that, these viewing angles for all testing 3D models are completely different from training pairs. Around 8k testing pairs are generated in total. This group of testing dataset is denoted as \textbf{cross viewing} (CV) angles testing dataset. Similarly, we also generate around 1.5k SV and 2.5k CV \textbf{validation data} split from another 12 CAD models, which are used for hyperparameter searching.

As our network is initially designed to predict an aligned full 3D model given a depth image from an arbitrary viewing angle, these two SV and CV testing datasets are generated separately to evaluate the viewing angle robustness and generality of our model.

Besides the large quantity of synthesized data, we also collect a \textbf{real-world dataset} in order to test the proposed network in a realistic scenario.
We use a Microsoft Kinect camera to manually scan a total of 20 object instances belonging to 4 classes \{bench, chair, couch, table\}, with 5 instances per class from different environments, including offices, homes, and outdoor university parks. For each object we acquire RGB-D images of the object from multiple angles by moving the camera around the object. Then, we use the dense visual SLAM algorithm ElasticFusion \cite{Whelan2015} in order to reconstruct the full 3D shape of each object, as well as the camera pose in each scan. 

We sample 50 random views from the camera trajectory, and for each one we obtain the depth image and the relative camera pose. In each depth image the 3D object is segmented from the background, using a combination of floor removal and manual segmentation. We finally generate ground truth information by aligning the full 3D objects with the partial 2.5D views.

\begin{figure}[t]
	\setlength{\abovecaptionskip}{ 0 cm}
   \setlength{\belowcaptionskip}{ -4pt}
   \centering
   \includegraphics[width=0.98\columnwidth]{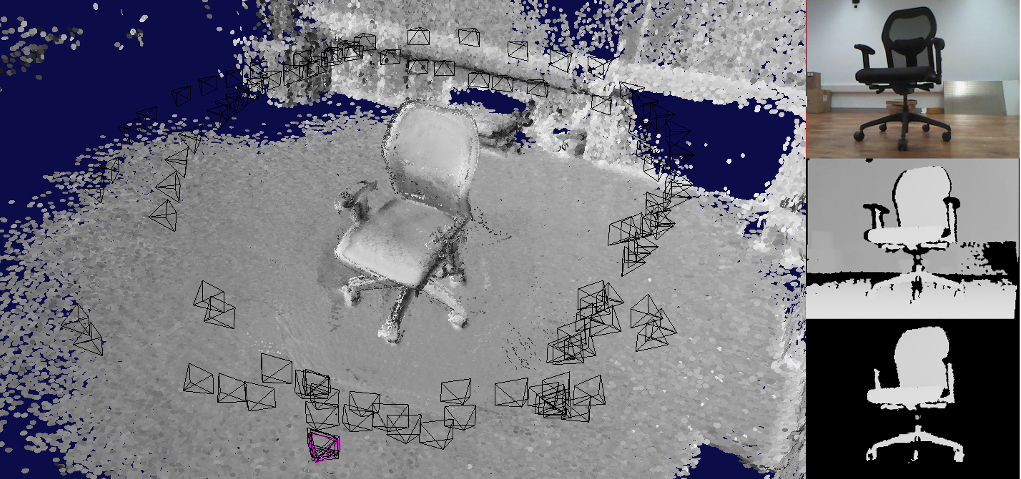}
   \caption{An example of ElasticFusion for generating real world data. Left: reconstructed object; sampled camera poses are shown in black. Right: Input RGB, depth image and segmented depth image.}
   \label{fig:elasticfusion}
   \vspace{-0.34 cm}
\end{figure}

It should be noted that, due to noise and quantization artifacts of low-cost RGB-D sensors, and the inaccuracy of the SLAM algorithm, the full 3D ground truth is not 100\% accurate, but can still be used as a reasonable approximation.
The real-world dataset highlights the challenges related to shape reconstruction from realistic data: noisy depth estimates, missing depth information, depth quantization. In addition, some of the objects are acquired outdoors (\eg{} $bench$), which is challenging for the near-infrared depth sensor of the Micorsoft Kinect.
However, we argue that a real-world benchmark for shape reconstruction is necessary for a thorough validation of future approaches. Figure \ref{fig:elasticfusion} shows an example of the reconstructed object and camera poses in ElasticFusion.

\section{Evaluation}
In this section, we evaluate our \nickname{} with comparison to the state of the art approaches and an ablation study to fully investigate the proposed network.

\subsection{Metrics}
To evaluate the performance of 3D reconstruction, we consider two metrics. The first metric is the mean Intersection-over-Union (IoU) between predicted 3D voxel grids and their ground truth. The IoU for an individual voxel grid is formally defined as follows:
\begin{equation*}
IoU = \frac{\sum_{i=1}^{N} \left[  I (y_i>p) * I(\bar{y_i}) \right] }{ \sum_{i=1}^{N}   \left[I  \left( I(y_{i} >p) + I(\bar{y_i}) \right) \right] } 
\end{equation*}
$where$ $I(\cdot)$ is an indicator function, $y_{i}$ is the predicted value for the $i^{th}$ voxel, $\bar{y_i}$ is the corresponding ground truth, $p$ is the threshold for voxelization, $N$ is the total number of voxels in a whole voxel grid. In all our experiments, $p$ is searched using the validation data split per category for each approach. Particularly, $p$ is searched in the range $[0.1, 0.9]$ with a step size $0.05$ using the validation datasets. The higher the IoU value, the better the reconstruction of a 3D model.

The second metric is the mean value of standard Cross-Entropy loss (CE) between a reconstructed shape and the ground truth 3D model. It is formally defined as: 
\begin{equation*}
CE =-\frac{1}{N} \sum_{i=1}^{N} \left[\bar{y_i}\log(y_i) + (1 - \bar{y_i})\log(1-y_i)\right]
\end{equation*}
$where$ $y_i$, $\bar{y_i}$ and $N$ are the same as defined in above IoU. The lower CE value is, the closer the prediction to be either `1' or `0', the more robust and confident the 3D predictions are. 

We also considered the Chamfer Distance (CD) or Earth Mover's Distance (EMD) as an additional metric. However, it is computationally heavy to calculate the distance between two high resolution voxel grids due to the large number of points. In our experiments, it takes nearly 2 minutes to calculate either CD or EMD between two $256^3$ shapes on a single Titan X GPU. Although the $256^3$ dense shapes can be downsampled to sparse point clouds on object surfaces to quickly compute CD or EMD, the geometric details are inevitably lost due to the extreme downsampling process. Therefore, we did not use CD or EMD for evaluation in our experiments.

\begin{figure*}[ht]
\setlength{\abovecaptionskip}{ 0 cm}
\setlength{\belowcaptionskip}{ 2 pt}
\centering
\begin{subfigure}[t]{0.98\textwidth}
   \includegraphics[width=1\linewidth]{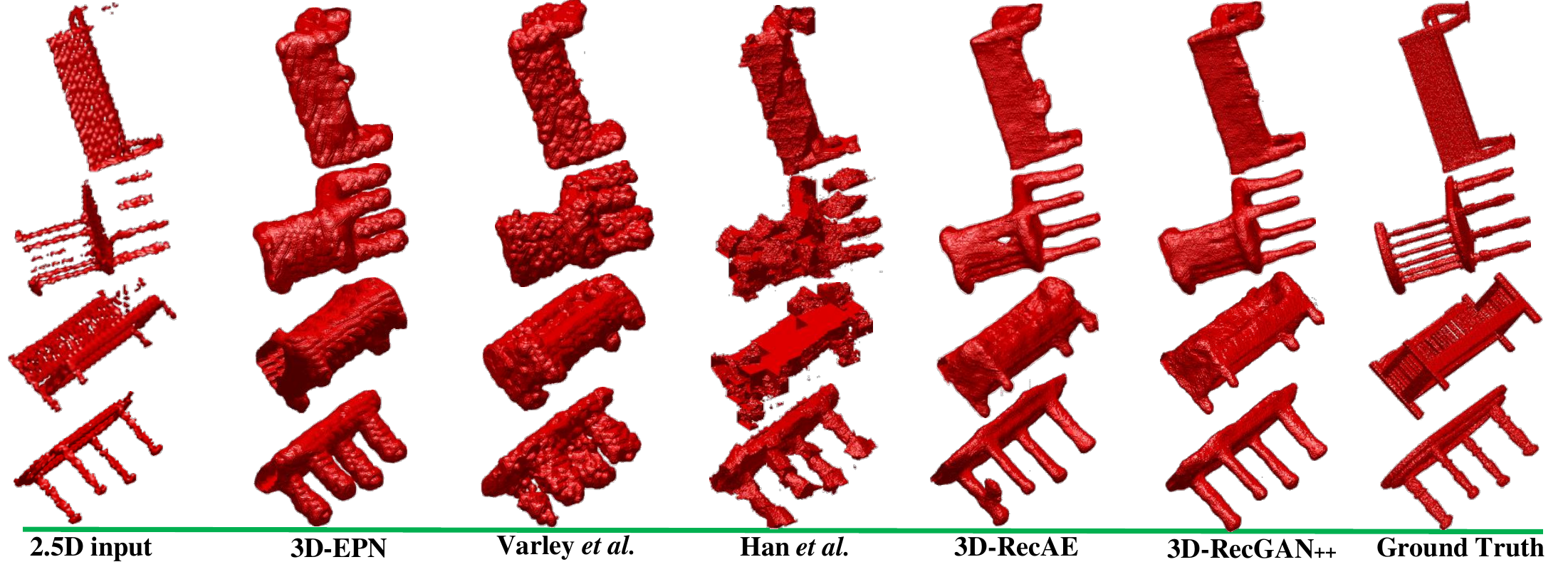}
   \caption{Qualitative results of single category reconstruction on testing datasets with same viewing angles.}
   \label{fig:per_cat_125} 
\end{subfigure}
\begin{subfigure}[t]{0.98\textwidth}
   \includegraphics[width=1\linewidth]{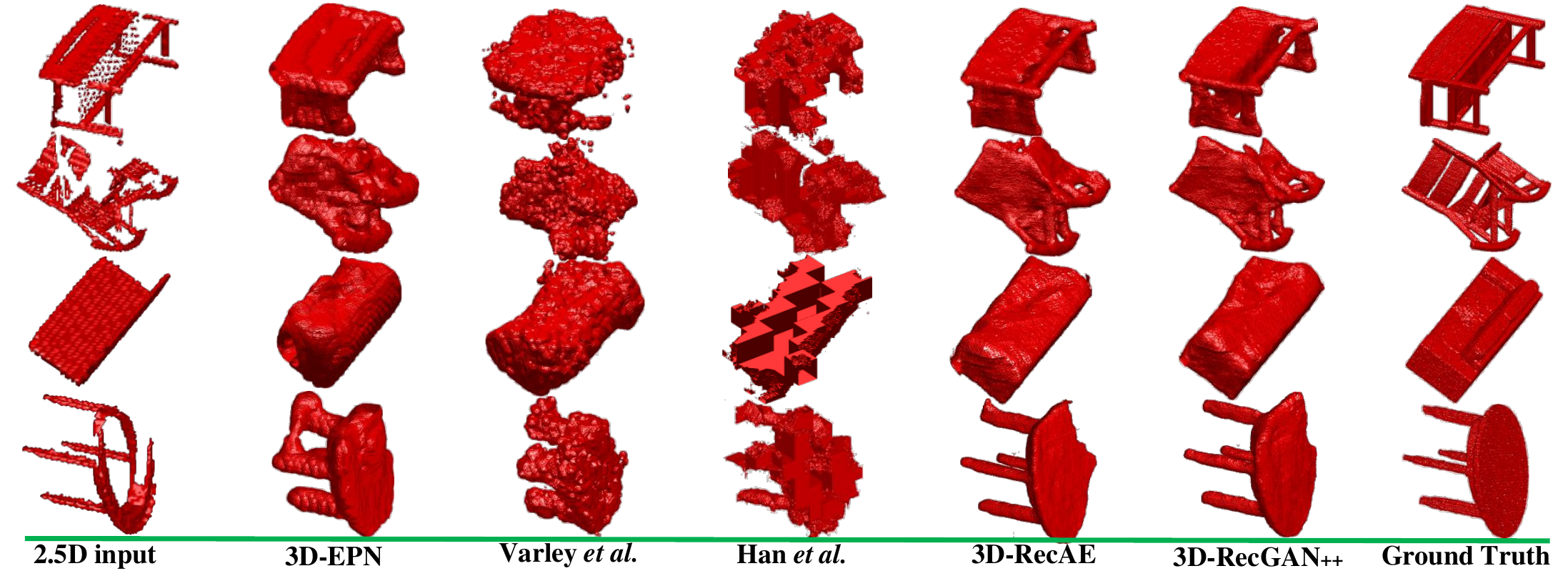}
   \caption{Qualitative results of single category reconstruction on testing datasets with cross viewing angles.}
   \label{fig:per_cat_216}
\end{subfigure}
\caption{Qualitative results of single category reconstruction on testing datasets with same and cross viewing angles.}
\label{fig:per_cat_125_216}
\vspace{-0.55 cm}
\end{figure*}

\subsection{Competing Approaches}

We compare against three state of the art deep learning based approaches for single depth view reconstruction. We also compare against the generator alone in our network, \ie{} without the GAN, named 3D-RecAE for short.

(1) \textbf{3D-EPN.} In \cite{Dai2017b}, Dai \etal{} proposed a neural network, called ``3D-EPN'', to reconstruct the 3D shape up to a $32^3$ voxel grid, after which a high resolution shape is retrieved from an existing 3D shape database, called ``Shape Synthesis''. In our experiment, we only compared with their neural network (\ie{} 3D-EPN) performance because we do not have an existing shape database for similar shape retrieval during testing. Besides, occupancy grid representation is used for the network training and testing.

(2) \textbf{Varley \etal{}} In \cite{Varley2017}, a network was designed to complete the 3D shape from a single 2.5D depth view for robot grasping. The output of their network is a $40^3$ voxel grid.

Note that, the low resolution voxel grids generated by 3D-EPN and Varley \etal{} are all upsampled to $256^3$ voxel grids using trilinear interpolation before calculating the IoU and CE metrics. The linear upsampling is a widely used post-processing technique for fair comparison in cases where the output resolution is not identical \cite{Tatarchenko2017}. However, as both 3D-EPN and Varley \etal{} are trained using lower resolution voxel grids for supervision, while the below Han \etal{} and our \nickname{} are trained using $256^3$ shapes for supervision, it is not strictly fair comparison in this regard. Considering both 3D-EPN and Varley \etal{} are among the early works and also solid competing approaches regarding the single depth view reconstruction task, we therefore include them as baselines.

(3) \textbf{Han \etal{}} In \cite{Han2017}, a global structure inference network and a local geometry refinement network are proposed to complete a high resolution shape from a noisy shape. The network is not originally designed for single depth view reconstruction, but its output shape is up to a $256^3$ voxel grid and is comparable to our network. For fair comparison, the same occupancy grid representation is used for their network. It should be noted that \cite{Han2017} involves convoluted designs, thus the training procedure is slower and less efficient due to many LSTMs integrated.

(4) \textbf{3D-RecAE.} As for our \nickname{}, we remove the discriminator and only keep the generator to infer the complete 3D shape from a single depth view. This comparison illustrates the benefits of adversarial learning.
\vspace{-0.08 cm}

\subsection{Single-category Results}
\begin{figure*}[ht]
\setlength{\abovecaptionskip}{ 0 cm}
\setlength{\belowcaptionskip}{ 2 pt}
\centering
\begin{subfigure}[t]{0.98\textwidth}
   \includegraphics[width=1\linewidth]{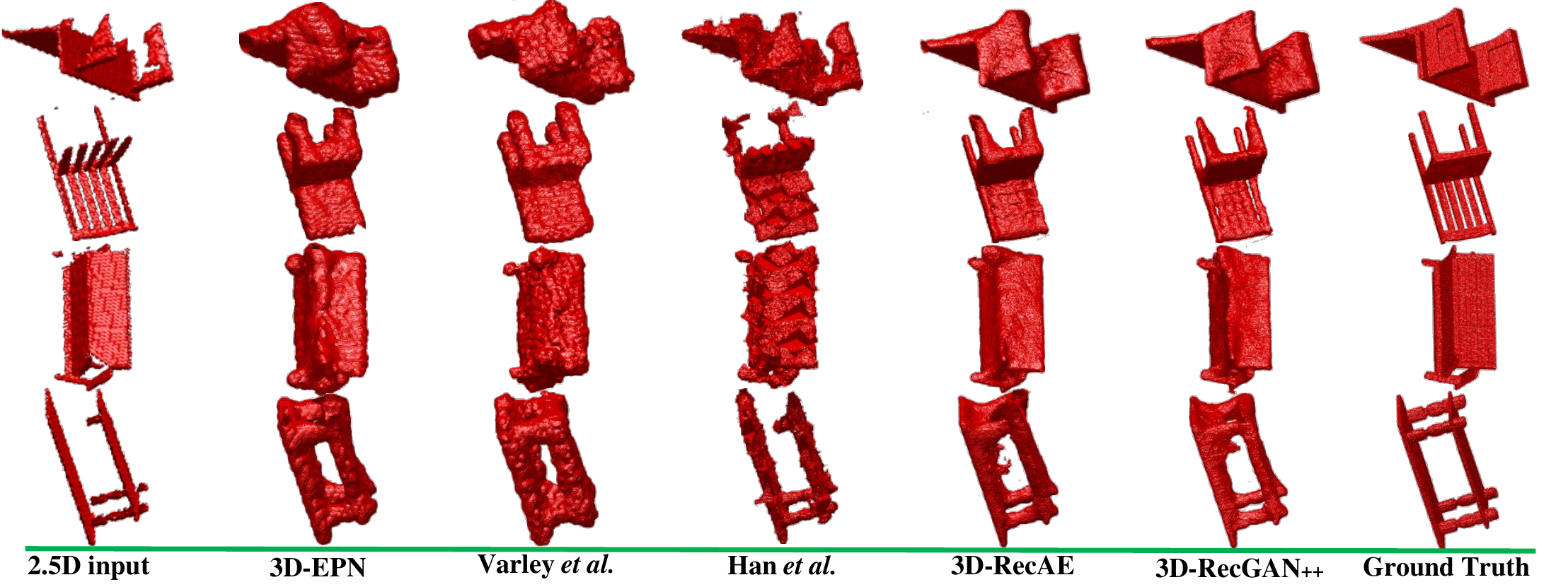}
   \caption{Qualitative results of multiple category reconstruction on testing datasets with same viewing angles.}
   \label{fig:multi_cat_125} 
\end{subfigure}
\begin{subfigure}[t]{0.98\textwidth}
   \includegraphics[width=1\linewidth]{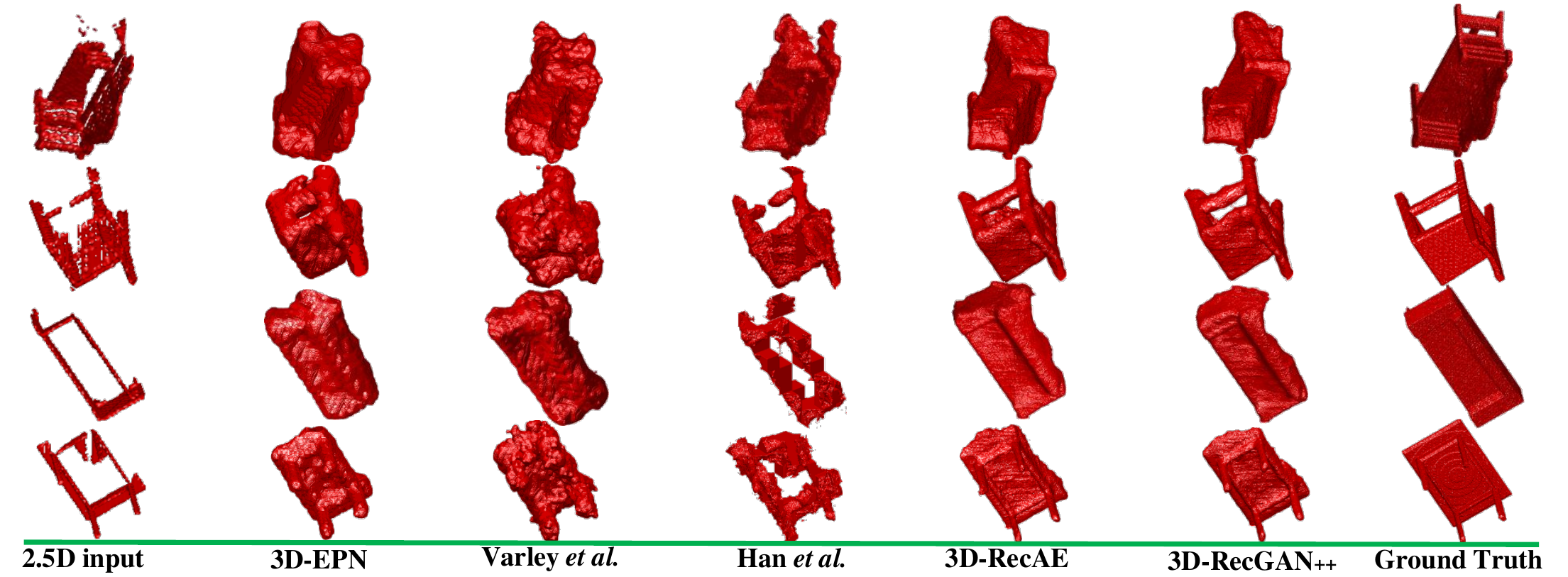}
   \caption{Qualitative results of multiple category reconstruction on testing datasets with cross viewing angles.}
   \label{fig:multi_cat_216}
\end{subfigure}
\caption{Qualitative results of multiple category reconstruction on testing datasets with same and cross viewing angles.}
\label{fig:multi_cat_125_216}
\vspace{-0.5 cm}
\end{figure*}

\hspace{1.5ex} (1) \textbf{Results.} All networks are separately trained and tested on four different categories, \{bench, chair, couch, table\}, with the same network configuration.  
Table \ref{tab:per_cat_iou_ce_sv_256} shows the IoU and CE loss of all methods on the testing dataset with same viewing angles on $256^3$ voxel grids, while Table \ref{tab:per_cat_iou_ce_cv_256} shows the IoU and CE loss comparison on testing dataset with cross viewing angles. Figure \ref{fig:per_cat_125_216} shows the qualitative results of single category reconstruction on testing datasets with same and cross viewing angles. The meshgrid function in Matlab is used to plot all 3D shapes for better visualization.

(2) \textbf{Analysis.} Both \nickname{} and 3D-RecAE significantly outperform the competing approaches in terms of IoU and CE loss on both the SV and CV testing datasets for dense 3D shape reconstruction ($256^3$ voxel grids). Although our approach is trained on depth input with a limited set of viewing angles, it still performs well to predict aligned 3D shapes from novel viewing angles. The 3D shapes generated by \nickname{} and 3D-RecAE are much more visually compelling than others.

Compared with 3D-RecAE, \nickname{} achieves better IoU scores and smaller CE loss. Basically, adversarial learning of the discriminator serves as a regularizer for fine-grained 3D shape estimation, which enables the output of \nickname{} to be more robust and confident. We also notice that the increase of \nickname{} in IoU and CE scores is not dramatic compared with 3D-RecAE. This is primarily because the main object shape can be reasonably predicted by 3D-RecAE, while the finer geometric details estimated by \nickname{} are usually smaller parts of the whole object shape. Therefore, \nickname{} only obtains a reasonable better IoU and CE scores than 3D-RecAE. The $4^{th}$ row of Figure \ref{fig:per_cat_125} shows a good example in terms of finer geometric details prediction of \nickname{}. In fact, in all the remaining experiments, \nickname{} is constantly, but not significantly, better than 3D-RecAE.

\vspace{-0.005 cm}
\begin{table}[h]
\caption{Per-category IoU and CE loss on testing dataset with same viewing angles ($256^3$ voxel grids).}
\vspace{-0.2 cm}
\centering
\label{tab:per_cat_iou_ce_sv_256}
\tabcolsep=0.05cm
\begin{tabular}{c|c|c|c|c|c|c|c|c}
	\hline
    	& \multicolumn{4}{c|}{IoU} & \multicolumn{4}{c}{CE Loss} \\ \hline
  	             & bench  & chair & couch & table & bench  & chair & couch & table \\ 
    \hlineB{3}
     3D-EPN \cite{Dai2017b}   & 0.423 & 0.488 & 0.631 & 0.508 &0.087 &0.105 &0.144 &0.101\\
    \hline
     Varley \etal{} \cite{Varley2017}   & 0.227 & 0.317 & 0.544 & 0.233 &0.111 &0.157 &0.195 &0.191\\
    \hline
     Han \etal{} \cite{Han2017}  & 0.441 & 0.426 & 0.446 & 0.499 & 0.045 & 0.081 &0.165 &0.058\\
    \hline
     \textbf{\scriptsize{3D-RecAE (ours)}}  & 0.577 & 0.641 & 0.749 & 0.675 &0.036 &0.063 &0.067 &0.043\\
     \hline
     \textbf{\scriptsize{\nickname{} (ours)}}  & \textbf{0.580} & \textbf{0.647} & \textbf{0.753} & \textbf{0.679} & \textbf{0.034} &\textbf{0.060}&\textbf{0.066} &\textbf{0.040} \\
\hline
\end{tabular}
\vspace{-0.3 cm}
\end{table}

\begin{table}[h]
\caption{Per-category IoU and CE loss on testing dataset with cross viewing angles ($256^3$ voxel grids).}
\vspace{-0.2 cm}
\centering
\label{tab:per_cat_iou_ce_cv_256}
\tabcolsep=0.05cm
\begin{tabular}{c|c|c|c|c|c|c|c|c}
	\hline
    	& \multicolumn{4}{c|}{IoU} & \multicolumn{4}{c}{CE Loss} \\ \hline
  	                          & bench  & chair & couch & table & bench  & chair & couch & table \\ 
    \hlineB{3}
     3D-EPN \cite{Dai2017b}   & 0.408 & 0.446 & 0.572 & 0.482 &0.086 &0.112 &0.163 &0.103\\
    \hline
     Varley \etal{} \cite{Varley2017}   & 0.185 & 0.278 & 0.475 & 0.187 &0.108 &0.171 &0.210 &0.186\\
    \hline
     Han \etal{} \cite{Han2017}  & 0.439 & 0.426 & 0.455 & 0.482 & 0.047 & 0.090 &0.163 &0.060\\
    \hline
     \textbf{\scriptsize{3D-RecAE (ours)}}  & 0.524 & 0.588 & 0.639 & 0.610 &0.045 &0.079 &0.117 &0.058\\
     \hline
     \textbf{\scriptsize{\nickname{} (ours)}}  & \textbf{0.531} & \textbf{0.594} & \textbf{0.646} & \textbf{0.618} & \textbf{0.041} &\textbf{0.074}&\textbf{0.111} &\textbf{0.053} \\
\hline
\end{tabular}
\vspace{-0.3 cm}
\end{table}


\subsection{Multi-category Results}\label{sec:multi_cat}
\vspace{-0.15 cm}
\hspace{1.5ex} (1) \textbf{Results.} All networks are also trained and tested on multiple categories without being given any class labels. The networks are trained on four categories: \{bench, chair, couch, table\}; and then tested separately on individual categories. Table \ref{tab:multi_cat_iou_ce_sv_256} shows the IoU and CE loss comparison of all methods on testing dataset with same viewing angles for dense shape reconstruction, while Table \ref{tab:multi_cat_iou_ce_cv_256} shows the IoU and CE loss comparison on testing dataset with cross viewing angles. Figure \ref{fig:multi_cat_125_216} shows the qualitative results of all approaches on testing datasets of multiple categories with same and cross viewing angles.

(2) \textbf{Analysis.} Both \nickname{} and 3D-RecAE significantly outperforms the state of the art by a large margin in all categories which are trained together on a single model. Besides, the performance of our network trained on multiple categories, does not notably degrade compared with training the network on individual categories as shown in previous Table \ref{tab:per_cat_iou_ce_sv_256} and \ref{tab:per_cat_iou_ce_cv_256}. This confirms that our network has enough capacity and capability to learn diverse features from multiple categories.

\begin{table}[h]
\caption{Multi-category IoU and CE loss on testing dataset with same viewing angles ($256^3$ voxel grids).}
\vspace{-0.2 cm}
\centering
\label{tab:multi_cat_iou_ce_sv_256}
\tabcolsep=0.05cm
\begin{tabular}{c|c|c|c|c|c|c|c|c}
	\hline
    	& \multicolumn{4}{c|}{IoU} & \multicolumn{4}{c}{CE Loss} \\ \hline
  	                          & bench  & chair & couch & table & bench  & chair & couch & table \\ 
    \hlineB{3}
     3D-EPN \cite{Dai2017b}   & 0.428 & 0.484 & 0.634 & 0.506 &0.087 &0.107 &0.138 &0.102\\
    \hline
     Varley \etal{} \cite{Varley2017}   & 0.234 & 0.317 & 0.543 & 0.236 &0.103 &0.132 &0.197 &0.170\\
    \hline
     Han \etal{} \cite{Han2017}  & 0.425 & 0.454 & 0.440 & 0.470 & 0.045 & 0.087 &0.172 &0.065\\
    \hline
     \textbf{\scriptsize{3D-RecAE (ours)}}  & 0.576 & 0.632 & 0.740 & 0.661 &0.037 &0.060 &0.069 &0.044\\
     \hline
     \textbf{\scriptsize{\nickname{} (ours)}}  & \textbf{0.581} & \textbf{0.640} & \textbf{0.745} & \textbf{0.667} & \textbf{0.030} &\textbf{0.051}&\textbf{0.063} &\textbf{0.039} \\
\hline
\end{tabular}
\vspace{-0.1 cm}
\end{table}

\begin{table}[h]
\caption{Multi-category IoU and CE loss on testing dataset with cross viewing angles ($256^3$ voxel grids).}
\vspace{-0.2 cm}
\centering
\label{tab:multi_cat_iou_ce_cv_256}
\tabcolsep=0.05cm
\begin{tabular}{c|c|c|c|c|c|c|c|c}
	\hline
    	& \multicolumn{4}{c|}{IoU} & \multicolumn{4}{c}{CE Loss} \\ \hline
  	                          & bench  & chair & couch & table & bench  & chair & couch & table \\ 
    \hlineB{3}
     3D-EPN \cite{Dai2017b}   & 0.415 & 0.452 & 0.531 & 0.477 &0.091 &0.115 &0.147 &0.111\\
    \hline
     Varley \etal{} \cite{Varley2017}   & 0.201 & 0.283 & 0.480 & 0.199 &0.105 &0.143 &0.207 &0.174\\
    \hline
     Han \etal{} \cite{Han2017}  & 0.429 & 0.444 & 0.447 & 0.474 & 0.045 & 0.089 &0.172 &0.063\\
    \hline
     \textbf{\scriptsize{3D-RecAE (ours)}}  & 0.530 & 0.587 & 0.640 & 0.610 &0.043 &0.068 &0.096 &0.055\\
     \hline
     \textbf{\scriptsize{\nickname{} (ours)}}  & \textbf{0.540} & \textbf{0.594} & \textbf{0.643} & \textbf{0.621} & \textbf{0.038} &\textbf{0.061}&\textbf{0.091} &\textbf{0.048} \\
\hline
\end{tabular}
\vspace{-0.45 cm}
\end{table}

\begin{figure*}[ht]
\setlength{\abovecaptionskip}{ 0 cm}
\setlength{\belowcaptionskip}{ 2 pt}
\centering
\begin{subfigure}[t]{0.98\textwidth}
   \includegraphics[width=1\linewidth]{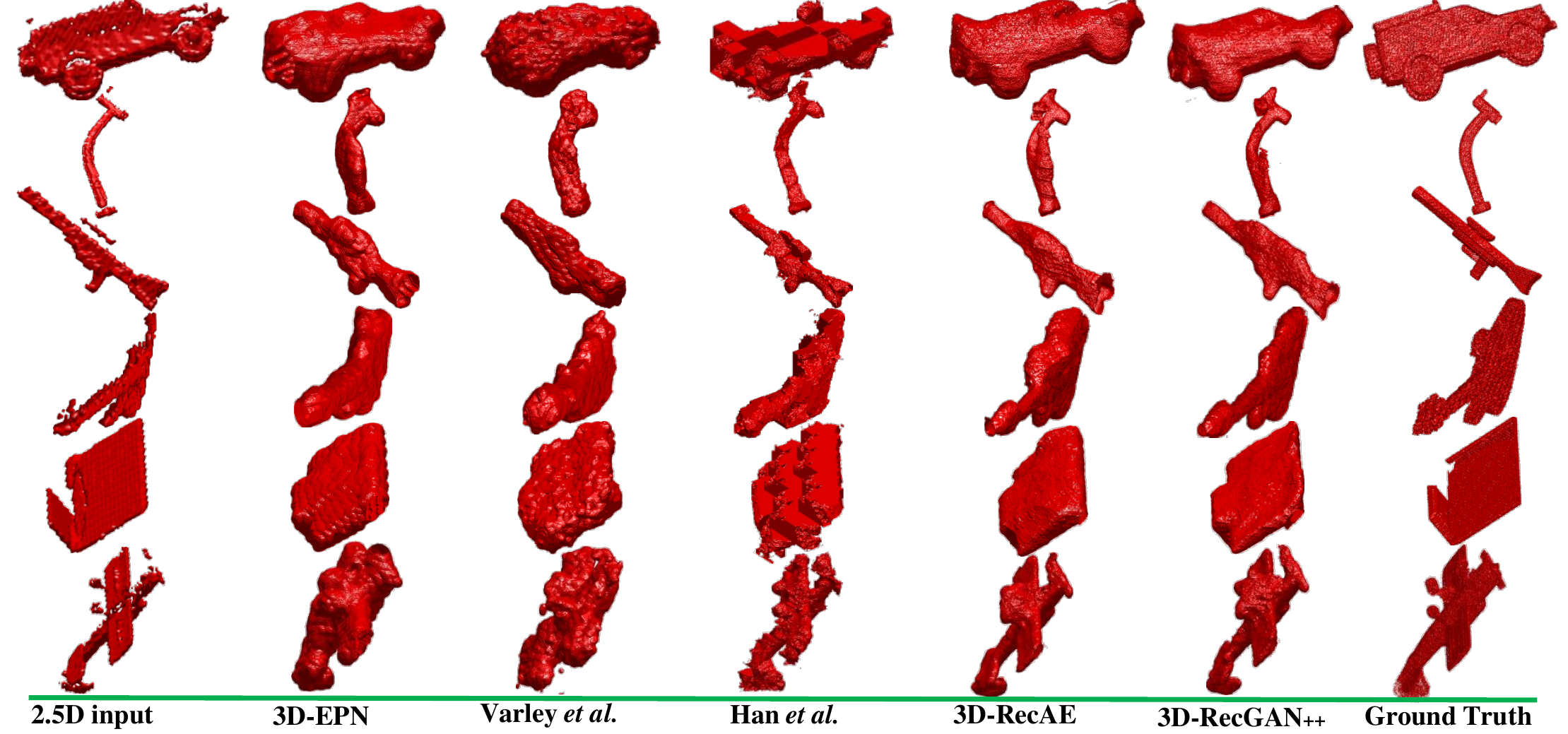}
   \caption{Qualitative results of cross category reconstruction on testing datasets with same viewing angles.}
   \label{fig:cross_cat_125} 
\end{subfigure}
\begin{subfigure}[t]{0.98\textwidth}
   \includegraphics[width=1\linewidth]{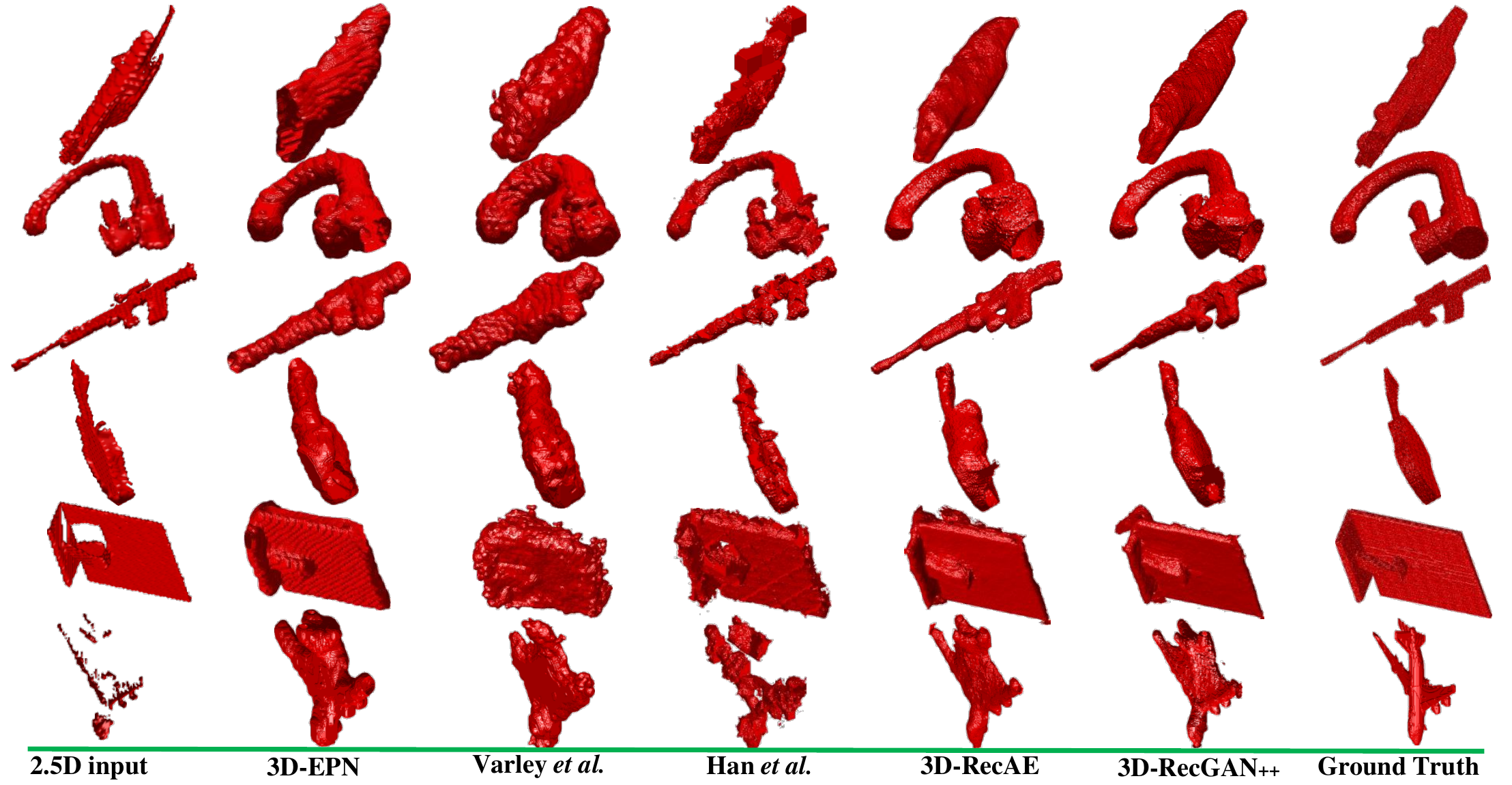}
   \caption{Qualitative results of cross category reconstruction on testing datasets with cross viewing angles.}
   \label{fig:cross_cat_216}
\end{subfigure}
\caption{Qualitative results of cross category reconstruction on testing datasets with same and cross viewing angles.}
\label{fig:cross_cat_125_216}
\vspace{-0.5 cm}
\end{figure*}
\vspace{-0.2 cm}
\subsection{Cross-category Results}
\begin{figure*}[t]
	\setlength{\abovecaptionskip}{ 0 cm}
    \setlength{\belowcaptionskip}{ -6pt}
    \centering
    \includegraphics[width=0.98\textwidth]{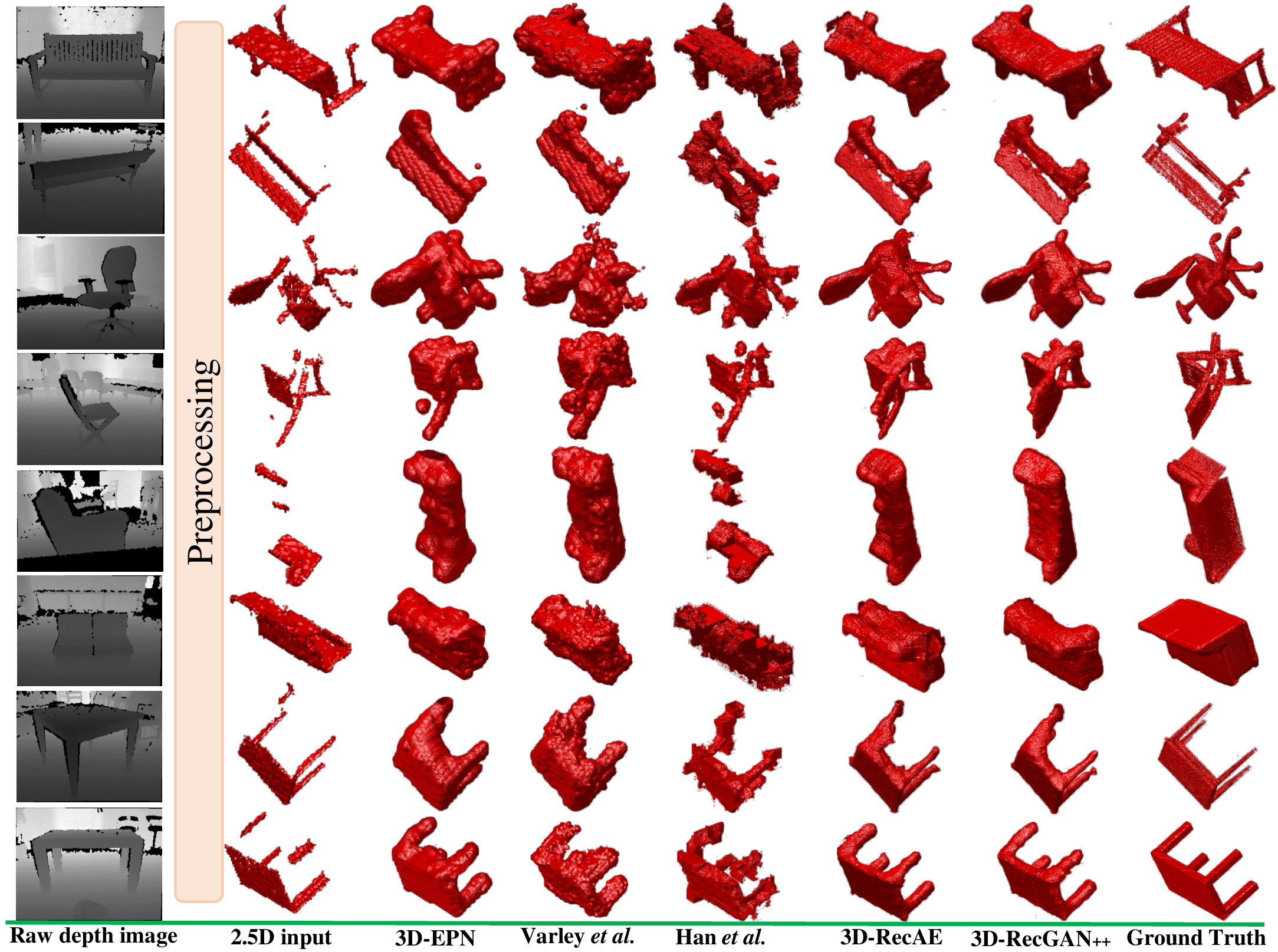}
    \caption{Qualitative results of real-world objects reconstruction from different approaches. The object instance is segmented from the raw depth image in preprocessing step.}
    \label{fig:multi_cat_real}
    \vspace{-0.1 cm}
\end{figure*}

\hspace{1.5ex} (1) \textbf{Results.} To further investigate the generality of networks, we train all networks on \{bench, chair, couch, table\}, and then test them on another 6 totally different categories: \{car, faucet, firearm, guitar, monitor, plane\}. For each of the 6 categories, we generate the same amount of testing datasets with same and cross viewing angles, which is similar to the previous \{bench, chair, couch, table\}. Table \ref{tab:cross_cat_iou_sv_256} and \ref{tab:cross_cat_ce_sv_256} shows the IoU and CE loss comparison of all approaches on the testing dataset with same viewing angles, while Table \ref{tab:cross_cat_iou_cv_256} and \ref{tab:cross_cat_ce_cv_256} shows the IoU and CE loss comparison on the testing dataset with cross viewing angles. Figure \ref{fig:cross_cat_125_216} shows the qualitative results of all methods on 6 unseen categories with same and cross viewing angles. 

\begin{table}[t]
\caption{Cross-category IoU on testing dataset with same viewing angles ($256^3$ voxel grids).}
\vspace{-0.2 cm}
\centering
\label{tab:cross_cat_iou_sv_256}
\tabcolsep=0.1cm
\begin{tabular}{c|c|c|c|c|c|c}
	\hline
  	 	 			 & car & faucet & firearm & guitar & monitor & plane \\ 
    \hlineB{3}
     3D-EPN \cite{Dai2017b}    & 0.450 & 0.442 & 0.339 & 0.351  & 0.444 & 0.314 \\
    \hline
     Varley \etal{} \cite{Varley2017}   & 0.484 & 0.260 & 0.280 & 0.255  & 0.341 & 0.295 \\
    \hline
     Han \etal{} \cite{Han2017}  & 0.360 & 0.402 & 0.333 & 0.353  & 0.450 & 0.306 \\
    \hline
     \textbf{\scriptsize{3D-RecAE (ours)}}  & \textbf{0.557} & 0.530 & 0.422 & 0.440  & 0.556 & 0.390 \\
    \hline
     \textbf{ \scriptsize{\nickname{} (ours)}}  & 0.555 & \textbf{0.536} & \textbf{0.426} & \textbf{0.442}  & \textbf{0.562} & \textbf{0.394} \\
\hline
\end{tabular}
\vspace{-0.3 cm}
\end{table}
\begin{table}[t]
\caption{Cross-category CE loss on testing dataset with same viewing angles ($256^3$ voxel grids).}
\vspace{-0.2 cm}
\centering
\label{tab:cross_cat_ce_sv_256}
\tabcolsep=0.1cm
\begin{tabular}{c|c|c|c|c|c|c}
	\hline
  	 	 	  & car & faucet & firearm & guitar & monitor & plane \\ 
    \hlineB{3}
     3D-EPN \cite{Dai2017b}    & 0.170 & 0.088 & 0.036 & 0.036  & 0.123 & 0.066 \\
    \hline
     Varley \etal{} \cite{Varley2017}   & 0.173 & 0.122 & 0.029 & 0.030  & 0.130 & 0.042 \\
    \hline
     Han \etal{} \cite{Han2017}  & 0.167 & 0.077 & 0.018 & 0.015  & 0.088 & 0.031 \\
    \hline
     \textbf{\scriptsize{3D-RecAE (ours)}}  & 0.110 & 0.057 & 0.018 & 0.016  & 0.072 & 0.036 \\
    \hline
     \textbf{ \scriptsize{\nickname{} (ours)}}  & \textbf{0.102} & \textbf{0.053} & \textbf{0.016} & \textbf{0.014}  & \textbf{0.067} & \textbf{0.031} \\
\hline
\end{tabular}
\vspace{-0.3 cm}
\end{table}
\begin{table}[t]
\caption{Cross-category IoU on testing dataset with cross viewing angles ($256^3$ voxel grids).}
\vspace{-0.2 cm}
\centering
\label{tab:cross_cat_iou_cv_256}
\tabcolsep=0.1cm
\begin{tabular}{c|c|c|c|c|c|c}
	\hline
  	 	 	  & car & faucet & firearm & guitar & monitor & plane \\ 
    \hlineB{3}
     3D-EPN \cite{Dai2017b}    & 0.446 & 0.439 & 0.324 & 0.359  & 0.448 & 0.309 \\
    \hline
     Varley \etal{} \cite{Varley2017}   & 0.489 & 0.260 & 0.274 & 0.255  & 0.334 & 0.283 \\
    \hline
     Han \etal{} \cite{Han2017}  & 0.349 & 0.402 & 0.321 & 0.363  & 0.455 & 0.299 \\
    \hline
     \textbf{\scriptsize{3D-RecAE (ours)}}  & 0.550 & 0.521 & 0.411 & 0.441  & 0.550 & 0.382 \\
    \hline
     \textbf{ \scriptsize{\nickname{} (ours)}}  & \textbf{0.553} & \textbf{0.529} & \textbf{0.416} & \textbf{0.449}  & \textbf{0.555} & \textbf{0.390} \\
\hline
\end{tabular}
\vspace{-0.3 cm}
\end{table}
\begin{table}[t]
\caption{Cross-category CE loss on testing dataset with cross viewing angles ($256^3$ voxel grids).}
\vspace{-0.2 cm}
\centering
\label{tab:cross_cat_ce_cv_256}
\tabcolsep=0.1cm
\begin{tabular}{c|c|c|c|c|c|c}
	\hline
  	 	 & car & faucet & firearm & guitar & monitor & plane \\ 
    \hlineB{3}
     3D-EPN \cite{Dai2017b}    & 0.160 & 0.087 & 0.033 & 0.036  & 0.127 & 0.065 \\
    \hline
     Varley \etal{} \cite{Varley2017}   & 0.171 & 0.123 & 0.028 & 0.030  & 0.136 & 0.043 \\
    \hline
     Han \etal{} \cite{Han2017}  & 0.171 & 0.076 & 0.018 & 0.016  & 0.088 & 0.031 \\
    \hline
     \textbf{\scriptsize{3D-RecAE (ours)}}  & 0.101 & 0.059 & 0.017 & 0.017  & 0.079 & 0.036 \\
    \hline
     \textbf{ \scriptsize{\nickname{} (ours)}}  & \textbf{0.100} & \textbf{0.055} & \textbf{0.014} & \textbf{0.015}  & \textbf{0.074} & \textbf{0.031} \\
\hline
\end{tabular}
\vspace{-0.3 cm}
\end{table}

We further evaluate the generality of our \nickname{} on a specific category. Particularly, we conduct four groups of experiments. In the first group, we train our \nickname{} on bench, then separately test it on the remaining 3 categories: \{chair, couch, table\}. In the second group, the network is trained on chair and separately tested on \{bench, couch, table\}. Similarly, another two groups of experiments are conducted. Basically, this experiment is to investigate how well our approach learns features from one category and then generalizes to a different category, and vice versa. Table \ref{tab:cross_cat_single_iou_ce_sv_256} shows the cross-category IoU and CE loss of our \nickname{} trained on individual category and then tested on the testing dataset with same viewing angles over $256^3$ voxel grids. 

(2) \textbf{Analysis.} The proposed \nickname{} achieves much higher IoU and smaller CE loss across the unseen categories than competing approaches. Our network not only learns rich features from different object categories, but also is able to generalize well to completely new types of categories. Our intuition is that the network may learn geometric features such as lines, planes, curves which are common across various object categories. As our network involves skip-connections between intermediate neural layers, it is not straightforward to visualize and analyze the learnt latent features.

It can be also observed that our model trained on $bench$ tends to be more general than others. Intuitively, the bench category tends to have general features such as four legs, seats, and/or a back, which are also common among other categories \{chair, couch, table\}. However, not all chairs or couches consist of such general features that are shared across different categories. 

Overall, we may safely conclude that the more similar features two categories share, including both the low-level lines/planes/curves and the high-level shape components, the better generalization of our model achieves cross those categories.

\begin{table}[h]
\caption{Cross-category IoU and CE loss of \nickname{} trained on individual category and then tested on the testing dataset with same viewing angles ($256^3$ voxel grids).}
\vspace{-0.2 cm}
\centering
\label{tab:cross_cat_single_iou_ce_sv_256}
\tabcolsep=0.07cm
\begin{tabular}{c|c|c|c|c|c|c|c|c}
	\hline
    	& \multicolumn{4}{c|}{IoU} & \multicolumn{4}{c}{CE Loss} \\ \hline
  	   & bench  & chair & couch & table & bench  & chair & couch & table \\ 
    \hlineB{3}
    \begin{tabular}{@{}c@{}} \scriptsize{\textbf{Group 1}} \\ \scriptsize{\textbf{(trained on bench)}}\end{tabular}  & \textbf{0.580} & \underline{0.510} & \underline{0.507} & \underline{0.599} &\textbf{0.034} &\underline{0.110} &\underline{0.164} &\underline{0.062}\\
    \hline
    \begin{tabular}{@{}c@{}}\scriptsize{Group 2} \\ \scriptsize{(trained on chair)}\end{tabular}  & 0.508 & \textbf{0.647} & 0.469 & 0.564 &0.048 &\textbf{0.060} &0.184 &0.069\\
    \hline
    \begin{tabular}{@{}c@{}}\scriptsize{Group 3} \\ \scriptsize{(trained on couch)}\end{tabular}  & 0.429 & 0.504 & \textbf{0.753} & 0.437 & 0.070 & 0.105 &\textbf{0.066} &0.126\\
    \hline
    \begin{tabular}{@{}c@{}}\scriptsize{Group 4} \\ \scriptsize{(trained on table)}\end{tabular}  & 0.510 & 0.509 & 0.402 & \textbf{0.679} &0.049 &0.111 &0.260 &\textbf{0.040}\\
\hline
\end{tabular}
\vspace{-0.5 cm}
\end{table}


\subsection{Real-world Experiment Results}
\hspace{1.5ex} (1) \textbf{Results.} Lastly, in order to evaluate the domain adaptation capability of the networks, we train all networks on synthesized data of categories \{bench, chair, couch, table\}, and then test them on real-world data collected by a Microsoft Kinect camera. Table \ref{tab:real_cat_iou_ce_256} compares the IoU and CE loss of all approaches on the real-world dataset. Figure \ref{fig:multi_cat_real} shows some qualitative results for all methods.

(2) \textbf{Analysis.} There are two reasons why the IoU is significantly lower compared with testing on the synthetic dataset. First, the ground truth objects obtained from ElasticFusion are not as solid as the synthesized datasets. However, all networks predict dense and solid voxel grids, so the interior parts may not match though the overall object shapes are satisfactorily recovered as shown in Figure \ref{fig:multi_cat_real}.
Secondly, the input 2.5D depth view from real-world dataset is noisy and incomplete, due to the limitation of the RGB-D sensor (\eg{} reflective surfaces, outdoor lighting). In some cases, the input 2.5D view does not capture the whole object and only contains a part of the object, which also leads to inferior reconstruction results (\eg{} the $5^{th}$ row in Figure \ref{fig:multi_cat_real}) and a lower IoU scores overall. However, our proposed network is still able to reconstruct reasonable 3D dense shapes given the noisy and incomplete 2.5D input depth views, while the competing algorithms (\eg{} Varley \etal{}) are not robust to real-world noise and unable to generate compelling results.

\begin{table}[h]
\caption{Multi-category IoU and CE loss on real-world dataset ($256^3$ voxel grids).}
\vspace{-0.2 cm}
\centering
\label{tab:real_cat_iou_ce_256}
\tabcolsep=0.05cm
\begin{tabular}{c|c|c|c|c|c|c|c|c}
	\hline
    	& \multicolumn{4}{c|}{IoU} & \multicolumn{4}{c}{CE Loss} \\ \hline
  	                          & bench  & chair & couch & table & bench  & chair & couch & table \\ 
    \hlineB{3}
     3D-EPN \cite{Dai2017b}   & 0.162 & 0.190 & 0.508 & 0.140 &0.090 &0.158 &0.413 &0.187\\
    \hline
     Varley \etal{} \cite{Varley2017}   & 0.118 & 0.152 & 0.433 & 0.075 &0.073 &0.155 &0.436 &0.191\\
    \hline
     Han \etal{} \cite{Han2017}  & 0.166 & 0.164 & 0.235 & 0.146 & 0.083 & 0.167 &0.352 &0.194\\
    \hline
     \textbf{\scriptsize{3D-RecAE (ours)}}  & 0.173 & 0.203 & 0.538 & 0.151 &0.065 &0.156 &0.318 &0.180\\
    \hline
     \textbf{\scriptsize{\nickname{} (ours)}}  & \textbf{0.177} & \textbf{0.208} & \textbf{0.540} & \textbf{0.156} & \textbf{0.061} &\textbf{0.153}&\textbf{0.314} &\textbf{0.177} \\
\hline
\end{tabular}
\vspace{-0.3 cm}
\end{table}

\clearpage

\subsection{Impact of Adversarial Learning}
\hspace{1.5ex} (1) \textbf{Results.} In all above experiments, the proposed \nickname{} tends to outperform the ablated network 3D-RecAE which does not include the adversarial learning of GAN part. In all visualization of experiment results, the 3D shapes from \nickname{} are also more compelling than 3D-RecAE. To further quantitatively investigate how the adversarial learning improves the final 3D results comparing with 3D-RecAE, we calculate the mean precision and recall from the previous multi-category experiment results in Section \ref{sec:multi_cat}. Table \ref{tab:multi_cat_pre_rec_sv_256} compares the mean precision and recall of \nickname{} and 3D-RecAE on individual categories using the network trained on multiple categories.

(2) \textbf{Analysis.} It can be seen that the results of \nickname{} tend to constantly have higher precision scores than 3D-RecAE, which means \nickname{} has less false positive estimations. Therefore, the estimated 3D shapes from 3D-RecAE are likely to be 'fatter' and 'bigger', while \nickname{} tends to predict 'thinner' shapes with much more shape details being exposed. 
Both \nickname{} and 3D-RecAE can achieve high recall scores (\ie{} above 0.8), which means both \nickname{} and 3D-RecAE are capable of estimating the major object shapes without too many false negatives. In other words, the ground truth 3D shape tends to be a subset of the estimated shape result.
\vspace{-0.15 cm}

\begin{table}[h]
\caption{Multi-category  mean precision and recall on testing dataset with same viewing angles ($256^3$ voxel grids).}
\vspace{-0.2 cm}
\centering
\label{tab:multi_cat_pre_rec_sv_256}
\tabcolsep=0.09cm
\begin{tabular}{c|c|c|c|c|c|c|c|c}
	\hline
    	& \multicolumn{4}{c|}{mean precision} & \multicolumn{4}{c}{mean recall} \\ \hline
  	          & bench  & chair & couch & table & bench  & chair & couch & table \\ 
    \hlineB{3}
     \scriptsize{3D-RecAE}  & 0.668 & 0.740 & 0.800 & 0.750 &\textbf{0.808} &0.818 &0.907 &0.845\\
     \hline
     \scriptsize{\nickname{}}  & \textbf{0.680} & \textbf{0.747} &\textbf{0.804} & \textbf{0.754} & 0.804 &\textbf{0.820}&\textbf{0.910} &\textbf{0.853} \\
\hline
\end{tabular}
\vspace{-0.23 cm}
\end{table}

Overall, with regard to experiments on per-category, multi-category, and cross-category experiments, our \nickname{} outperforms others by a large margin, although all other approaches can reconstruct reasonable shapes. In terms of the generality, Varley \etal{} \cite{Varley2017} and  Han \etal{} \cite{Han2017} are inferior because Varley \etal{} \cite{Varley2017} use a single fully connected layers, instead of 3D ConvNets, for shape generation which is unlikely to be general for various shapes, and Han \etal{} \cite{Han2017} apply LSTMs for shape blocks generation which is inefficient and unable to learn general 3D structures. However, our \nickname{} is superior thanks to the generality of the 3D encoder-decoder and the adversarial discriminator. Besides, the 3D-RecAE tends to over estimate the 3D shape, while the adversarial learning of \nickname{} is likely to remove the over-estimated parts, so as to leave the estimated shape to be clearer with more shape details.
\vspace{-0.3 cm}

\subsection{Computation Analysis}
Table \ref{tab:para_time} compares the computation efficiency of all approaches regarding the total number of model parameters and the average time consumption to recover a single object. 

The model proposed by Han \etal{} \cite{Han2017} has the least number of parameters because most of the parameters are shared to predict different blocks of an object. Our \nickname{} has reasonable 167.1 millions parameters, which is on a similar scale to VGG-19 (\ie{} 144 millions) \cite{Simonyan2015}.

To evaluate the average time consumption for a single object reconstruction, we implement all networks in Tensorflow 1.2 and Python 2.7 with CUDA 8.0 and cuDNN 7.1 as the back-end driver and library. All models are tested on a single Titan X GPU in the same hardware and software environments. 3D-EPN \cite{Dai2017b} takes the shortest time to predict a $32^3$ object on GPU, while our \nickname{} only needs around 40 milliseconds to recover a dense $256^3$ object. Comparatively, Han \etal{} takes the longest GPU time to generate a dense object because of the time-consuming sequential processing of LSTMs. The low resolution objects predicted by 3D-EPN and Varley \etal{} are further upsampled to $256^3$ using existing SciPy library on a CPU server (Intel E5-2620 v4, 32 cores). It takes around 7 seconds to finish the upsampling for a single object.
\vspace{-0.15 cm}

\begin{table}[h]
\caption{Comparison of model parameters and average time consumption to reconstruction a single object.}
\vspace{-0.3 cm}
\centering
\label{tab:para_time}
\tabcolsep=0.165cm
\begin{tabular}{c|c|c|c}
	\hline
    	& \multicolumn{1}{c|}{\begin{tabular}{@{}c@{}} \textbf{parameters} \\ (millions) \end{tabular}} & \multicolumn{1}{c|}{ \begin{tabular}{@{}c@{}} \textbf{GPU time} \\ \scriptsize{(milliseconds)} \end{tabular} }  &  \multicolumn{1}{c}{ \begin{tabular}{@{}c@{}} \textbf{predicted 3D shapes} \\ \scriptsize{(resolution)} \end{tabular} }\\ 
    \hlineB{3}
    3D-EPN \cite{Dai2017b}   & 52.4 & \textbf{15.8} & $32^3$ \\ 
    \hline
    Varley \etal{} \cite{Varley2017} & 430.3 & 16.1 & $40^3$\\
    \hline
    Han \etal{} \cite{Han2017} & \textbf{7.5} & 276.4 &\boldsymbol {$256^3$}\\
    \hline
    \scriptsize{\nickname{}}  & 167.1 & 38.9 & \boldsymbol{$256^3$} \\
\hline
\end{tabular}
\vspace{-0.6 cm}
\end{table}

\section{Discussion}
Although our \nickname{} achieves the state of the art performance in 3D object reconstruction from a single depth view, it has limitations. Firstly, our network takes the volumetric representation of a single depth view as input, instead of taking a raw depth image. Therefore, a preprocessing of raw depth images is required for our network. However, in many application scenarios such as robot grasping, such preprocessing would be trivial and straightforward given the depth camera parameters. Secondly, the input depth view of our network only contains a clean object information without cluttered background. One possible solution is to leverage an existing segmentation algorithm such as Mask-RCNN \cite{He2017a} to clearly segment the target object instance from the raw depth view.

\section{Conclusion}
In this work, we proposed a framework \nickname{} that reconstructs the full 3D structure of an object from an arbitrary depth view. By leveraging the generalization capabilities of 3D encoder-decoder and generative adversarial networks, our \nickname{} predicts dense and accurate 3D structures with fine details, outperforming the state of the art in single-view shape completion for individual object category. We further tested our network's ability to reconstruct multiple categories without providing any object class labels during training or testing, and it showed that our network is still able to predict precise 3D shapes. Besides, we investigated the network's reconstruction performance on unseen categories, our proposed approach can also predict satisfactory 3D structures. Finally, our model is robust to real-world noisy data and can infer accurate 3D shapes although the model is purely trained on synthesized data. This confirms that our network has the capability of learning general 3D latent features of the objects, rather than simply fitting a function for the training datasets, and the adversarial learning of \nickname{} learns to add geometric details for estimated 3D shapes. In summary, our network only requires a single depth view to recover a dense and complete 3D shape with fine details.
\ifCLASSOPTIONcaptionsoff
  \newpage
\fi

\clearpage
\bibliographystyle{IEEEtran}
\bibliography{Mendeley}
\vspace{-4cm}
\begin{IEEEbiography}[{\includegraphics[width=1in,height=1.25in,clip,keepaspectratio]{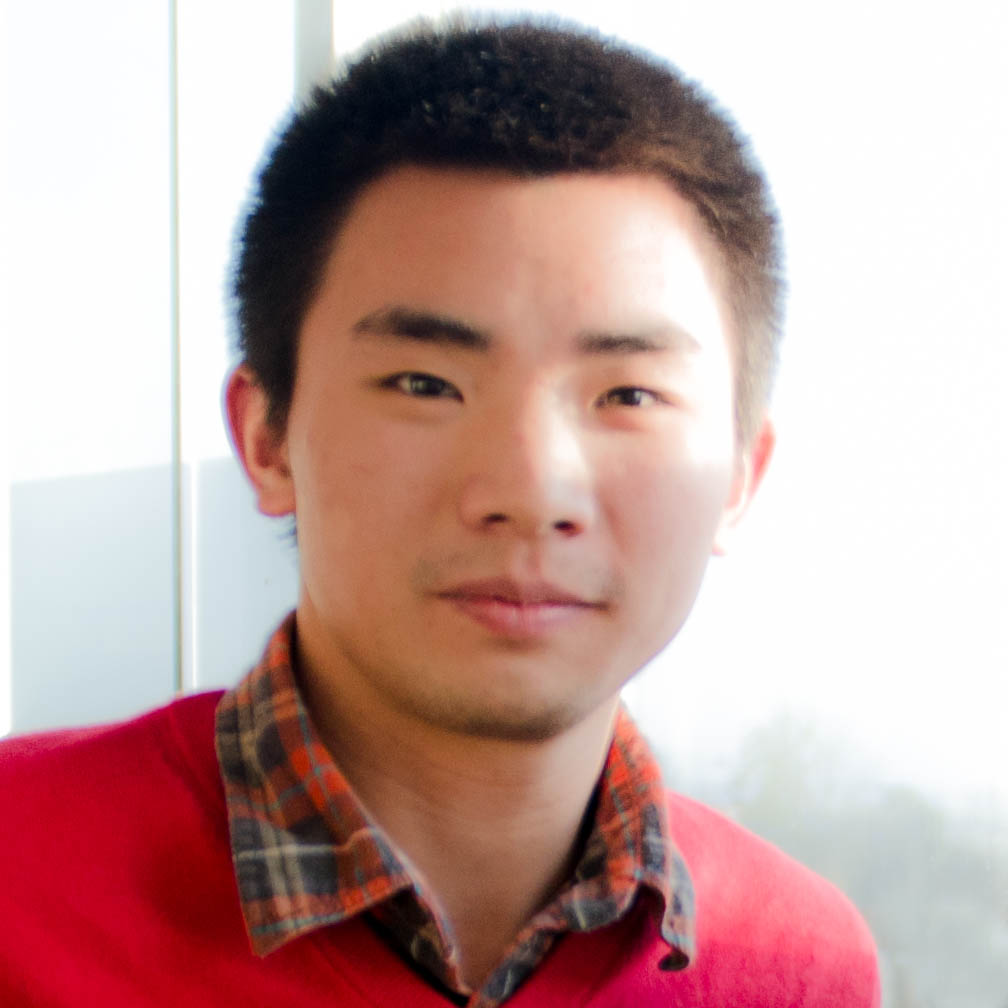}}]{Bo Yang} is currently a DPhil candidate in the department of Computer Science at University of Oxford. His research interests lie in deep learning, computer vision and robotics.\end{IEEEbiography}

\vspace{-4.3cm}
\begin{IEEEbiography}[{\includegraphics[width=1in,height=1.25in,clip,keepaspectratio]{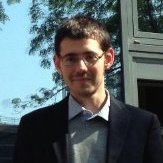}}]{Stefano Rosa} received his Ph.D. in Mechatronics Engineering from Politecnico di Torino, Italy, in 2014. He is currently a research fellow in the Department of Computer Science at University of Oxford, UK, working on long-term navigation, Human-Robot Interaction and intuitive physics.\end{IEEEbiography}

\vspace{-4.3cm}
\begin{IEEEbiography}[{\includegraphics[width=1in,height=1.25in,clip,keepaspectratio]{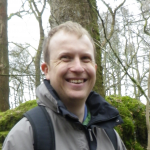}}]{Andrew Markham} is an Associate Professor at the Department of Computer Science, University of Oxford. He obtained his BSc (2004) and PhD (2008) degrees from the University of Cape Town, South Africa. He is the Director of the MSc in Software Engineering. He works on resource-constrained systems, positioning systems, in particular magneto-inductive positioning and machine intelligence.\end{IEEEbiography}

\vspace{-4.3cm}
\begin{IEEEbiography} [{\includegraphics[width=1in,height=1.25in,clip,keepaspectratio]{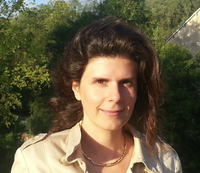}}]{Niki Trigoni} is a Professor at the Oxford University Department of Computer Science and a fellow of Kellogg College. She obtained her DPhil at the University of Cambridge (2001), became a postdoctoral researcher at Cornell University (2002-2004), and a Lecturer at Birkbeck College (2004-2007). At Oxford, she is currently Director of the EPSRC Centre for Doctoral Training on Autonomous Intelligent Machines and Systems, a program that combines machine learning, robotics, sensor systems and verification/control. She also leads the Cyber Physical Systems Group http://www.cs.ox.ac.uk/activities/sensors/index.html, which is focusing on intelligent and autonomous sensor systems with applications in positioning, healthcare, environmental monitoring and smart cities. The group’s research ranges from novel sensor modalities and low level signal processing to high level inference and learning. \end{IEEEbiography}

\vspace{-3.96cm}
\begin{IEEEbiography}[{\includegraphics[width=1in,height=1.25in,clip,keepaspectratio]{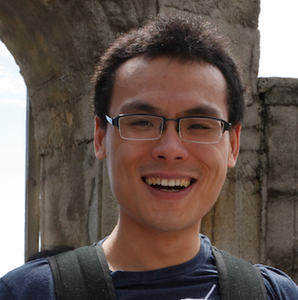}}]{Hongkai Wen} is an Assistant Professor at the Department of Computer Science, University of Warwick. Before that he was a post-doctoral researcher in the University of Oxford, where he obtained his D.Phil in Computer Science in 2014. His research interests lie in Cyber-Physical systems, human-centric sensing, and pervasive data science. \end{IEEEbiography}

\end{document}